\documentclass{article}
\usepackage{graphicx} 
\usepackage{url}

\usepackage{times}
\usepackage{latexsym}

\usepackage[many]{tcolorbox}  
\newtcolorbox{boxRound}{
    boxrule = 1.5pt,
    rounded corners,
    arc = 5pt   
}

\usepackage{graphicx}

\usepackage{booktabs}       
\usepackage{amsfonts}
\usepackage{nicefrac}       
\usepackage{microtype}      
\usepackage{lipsum}         
\usepackage{wrapfig}
\usepackage{doi}
\usepackage{amsmath}
\usepackage{amssymb}
\usepackage{amsthm}
\usepackage{bbding}
\usepackage{multirow}
\usepackage{subcaption}
\usepackage{algorithm}
\usepackage{algorithmic}
\usepackage{enumitem}
\usepackage{makecell}       
\usepackage{color}
\usepackage{hyperref}       
\usepackage{minitoc}
\usepackage{adjustbox}

\usepackage{array} %
\usepackage{colortbl}
\usepackage{pifont}%

\newcolumntype{a}{>{\columncolor{gray!10}}c}

\usepackage[preprint]{neurips_style_files/neurips_2020}
\title{Behemoth: Benchmarking Unlearning in LLMs Using Fully Synthetic Data}
\author{Eugenia Iofinova, Dan Alistarh\\IST Austria }

\begin{document}

\maketitle

\begin{abstract}
    As artificial neural networks, and specifically large language models, have improved rapidly in capabilities and quality, they have increasingly been deployed in real-world applications, from customer service to Google search, despite the fact that they frequently make factually incorrect or undesirable statements. This trend has inspired practical and academic interest in model editing, that is, in adjusting the weights of the model to modify its likely outputs for queries relating to a specific fact or set of facts. This may be done either to amend a fact or set of facts, for instance, to fix a frequent error in the training data, or to suppress a fact or set of facts entirely, for instance, in case of dangerous knowledge. Multiple methods have been proposed to do such edits. However, at the same time, it has been shown that such model editing can be brittle and incomplete. 
    However, the effectiveness of any model editing method necessarily depends on the data on which the model is trained, and, therefore, a good understanding of the interaction of the training data distribution and the way it is stored in the network is necessary and helpful to reliably perform model editing. However, working with large language models trained on real-world data does not allow us to understand this relationship or fully measure the effects of model editing. We therefore propose Behemoth, a fully synthetic data generation framework. To demonstrate the practical insights from the framework, we explore model editing in the context of simple tabular data, demonstrating surprising findings that, in some cases, echo real-world results, for instance, that in some cases restricting the update rank results in a more effective update. The code is available at \url{https://github.com/IST-DASLab/behemoth.git}.
\end{abstract}

\section{Introduction}

 Knowledge editing in large language models (LLMs) has acquired increasing practical importance as the scope of their applications has grown. However, despite substantial practical and academic interest, it remains largely an unsolved problem, as the changes have been shown to be brittle~\citep{Lynch2024EightMethodsRobustUnlearning} and also to have the potential to affect model accuracy in other areas. However, model unlearning evaluation methods typically focus on LLMs trained on large natural language datasets, which makes it necessary to rely on imperfect proxies, both to test the extent of the unlearning and to evaluate the impact on the rest of the model. Additionally, the very large natural-language datasets combined with the post-training alignment step make it difficult to investigate the relationship between characteristics of the training data and how well the knowledge in the model may be edited. Thus, we propose to investigate this in the realm of \emph{fully synthetic} data, with a fully specified data distribution and sentence grammar. To this end, we present a library that generates tabular data in the form of tuples of \{subject, relationship, object \}. These tuples are then combined into sentences using one or several pre-determined, artificial grammars. We extend earlier works \citep{allenzhu2023physics, Wu2024EvaluatingDU}, by implementing a completely custom vocabulary and tokenization, in which each token serves a specific function: either belongs uniquely to the subject, relationship, or object space of tokens, or has a unique grammatical function, for instance the \texttt{endofstring} token, and by more thoroughly investigating knowledge editing in this context, focusing on low-rank knowledge editing, on various knowledge editing desiderata, and on the scale of the edit. (While we believe that token collisions are interesting to study in their own right, we propose that any such token overuse be created intentionally and leave this investigation to future work.)

 We then use these data to train GPT-style Pythia Transformer architectures of size 31 million parameters. This size was chosen as the largest with which we could feasibly run a large number of experiments.

We then evaluate whether, after training, a particular piece of information can be edited and what the exact consequences of that editing are on the rest of the model. This improves over the use of proxy measurements, such as LLM benchmarks or hand-curated validation sets often collected with other LLMs to estimate whether a specific fact was fully edited by the model, and measure any collateral model damage. It also improves on previous works such as \cite{Maini2024TOFU} that insert information into the model using additional training (fine-tuning) of pre-trained LLMs, which may have different data storage implications than if the information were presented early on in training (nor does it allow for a custom tokenization, which prevents token collisions across unrelated concepts). In other words, the experiments presented in this paper may be thought of as a tradeoff of realism, in the sense that the data used in these experiments only loosely resemble real-world data, for measurement precision.

Overall, our framework provides a novel way to study how models store data, as well as more precisely benchmark model editing algorithms and the differences between them, while the choice of a small architecture and dataset allows many more experiments to be run than when using larger models that can reasonably hold natural-language knowledge. Already, even in our simple setup, we demonstrate notable differences between the action of full finetuning, low-rank finetuning, and ROME model editing, as well as a study of the impact of the choice of model layers to edit.
Further, while we apply this library to investigating knowledge editing, we note that it can easily be extended to investigating how other aspects of LLMs, such as model compression or scaling laws, depend on the type of data stored in the model.

\section{Approach}

The main idea of the synthetic data framework is to generate a set of \textit{facts}, which in our conception consist of a \emph{subject}, a \emph{relationship}, and an \emph{object} - which could be envisioned as corresponding to factual information such as \{\textit{Germany, Chancellor, Friedrich Merz}\}. These facts are then generated according to a pre-set underlying distribution, resulting in a set of \{\textit{subject, relationship, object}\} (hence, $\{s, o, r\}$) tuples. These tuples are then arranged into sentences using a predefined grammar, and these sentences form the training data of the model. In practice, once an architecture is selected, the number of tuples is chosen so that the model performance is high but not perfect, as described in the next section. 

Unlike other works, such as~\cite{allenzhu2023physics}, we use a fully synthetic vocabulary and a custom tokenization, with all tokens fully partitioned into subjects, objects, relationships, and different grammatical roles, such as the \texttt{<|endofstring|>} token. This allows us to fully control not only the fact, but also the token distribution, and also allows for additional quality checks as the models are trained and edited, as we can easily and explicitly check both the facts and the grammar. The tokenization is computed automatically by the Behemoth framework once the key parameters of the data and grammar are set.

The framework is designed to be modular, so that fact generation and sentence construction can be manipulated separately, and, in fact, the framework comes with several options for fact generation and for sentence construction (including various approaches to sentence formation), and can easily be expanded in the future.

While we acknowledge that synthetic data, such as that created by Behemoth, cannot reasonably be expected to rival the complexity of real natural-language data, and that tabular data is just one type of data contained in natural language models, we believe that the framework is helpful for studying LLMs and LLM editing for the following reasons. First, a large portion of model editing desiderata specifically concerns \emph{factual data} not dissimilar to that embodied by the $\{s, o, r\}$ tuples in Behemoth, and thus the task of editing one or several tuples, which we explore in this paper, parallels data editing required for fact correction in real LLMs. Second, works such as ours draw attention to the role of data and the data distribution in the effectiveness of model editing methods. Specifically, the true target and output data distribution are not known for natural language models (especially after alignment steps), which makes changes in the output data distribution post-editing difficult to measure. While the current work only scratches the surface of what is possible with synthetic data, we hope that releasing this framework will encourage additional exploration in this space.

\section{Related Work}
Knowledge unlearning and editing methods have received substantial interest since the rise of large language models, largely due to the practical necessity of such methods if LLMs are to be deployed for important real-world tasks. Thus, unlearning and knowledge editing have been studied both in LLMs trained on real data, and, to a much smaller extent, in LLMs trained on real data and fine-tuned on synthetic natural language data, such as \cite{Maini2024TOFU}. By contrast, this work studies knowledge editing in LLMs trained entirely on synthetic data.

This work was heavily influenced by the Physics of Language Modeling series, specifically \cite{ allenzhu2023physics}. In this work, the authors use fake biographies to study the necessary conditions for effective data retrieval by means of plain English question-answering. They find that data augmentations are essential to allow the extension of the task (question answering) to rely on stored knowledge rather than question memorization. This paper is part of a series that also includes \cite{allenzhu2024physics1} and \cite{allenzhu2024physics}, which use a synthetic syntactic language to train models to test strings for syntactic correctness, and examine the ability of transformers to manipulate information in inferring implied relationships, and finally measure the information storage of LLMs to derive a version of scaling laws. However, these works use real language in their synthetic data, which reduces some amount of control over the dataset tokenization and does not prevent collisions. \cite{Krishnan2025NotAD} uses the synthetic data framework developed in these works and unlearning via gradient ascent to show that more frequent instances in the training data are harder to unlearn and damage the resulting model.

Multiple works have found model editing, and, in particular, fine-tuning, to be brittle, among them \citep{Qi2024GPTJailbreak}, which focuses on breaking alignment fine-tuning in Llama2-7B-Chat by finding the neurons and, alternatively, subspaces of weight matrices that are most responsible for alignment via SNIP~\citep{lee2018snip} and Wanda~\citep{Sun2023Wanda} and then modifying these weights. In the follow-up paper \cite{qi2024safetyalignment}, the authors show that safety alignment fine-tuning primarily affects the first few tokens of prompts and demonstrate that this leads to a host of attacks. Similarly, ~\cite{jain2023mechanistically} and \cite{Wei2024Brittle} demonstrate that alignment finetuning learns a fairly narrow wrapper over existing functionality that can easily be undone, mostly in a synthetic setting. The overall conclusion is similar to~\cite{Qi2024GPTJailbreak}, in that the authors call out safety alignment as a specific case of easy unlearning.

Additionally, many papers have focused on designing or comparing various types of model editing. Famously, in~\citep{eldan2023whosHarryPotter} the authors largely unlearn all Harry Potter knowledge by finetuning the model on alternative data that essentially replaces the correct information with other information. However, in~\citep{Lynch2024EightMethodsRobustUnlearning}, a closer audit of the No-Harry model shows that much of the knowledge is still retained in the original model, if creatively retrieved. In Locating and Editing Factual Associations in GPT~\citep{meng2022locating}, the authors use a causal approach to find the neuron activations responsible for storing specific facts, in the style of $\{s, o, r\}$ tuples. They show that they can find the neuron activations in MLP layers and edit them to change the stored knowledge, on par with other methods such as finetuning. The follow-up work ~\citep{meng2023memit} proposes an algorithm for making a large number of such edits at the same time.
In Lora Learns Less and Forgets Less~\citep{Biderman2024lora}, the authors have found that LoRA underperforms FFT on more complex tasks, such as code generation, but also does a better job maintaining other skills and output diversity. Further, they show that FFT is not low-rank for more complex tasks. \cite{kotha2024understandingCatastrophicForgetting} shows that finetuning often `unlearns' on the level of task recognition, rather than actually forgetting knowledge. For instance, they recover ``unlearned'' knowledge by translating the prompts into a different language. In \cite{Hong2024DissectingFU}, the authors analyze fine-tuning for unlearning on real data and find that frequently the fine-tuning simply suppresses the output in the last layer. In \cite{Joshi2024TowardsRE}, the authors use data transformations to evaluate unlearning by checking if changing the query returns the desired response. In \citep{Wu2024EvaluatingDU}, the authors fine-tune models with additional facts with deep relationships, and find that various unlearning methods either fail to unlearn implied facts or destroy other unrelated information. Finally, \cite{hartmann2023sok} proposes a taxonomy (more like a pyramid) of different types of memorization and catalogs some mitigations, and some open areas.

Another important direction is the use of mechanical interpretability to identify and manipulate functionality in neural networks. In ~\cite{Bricken2024ToyMonosemanticity} and \cite{Templeton2024ClaudeMonosemanticity}, the authors use dictionary learning, in the form of a sparse autoencoder, to ascribe linear combinations of neuron activations to concepts in the data. Then, in~\citep{Templeton2024ClaudeMonosemanticity}, the researchers show that they can use these 'features' to manipulate the model by synthetically manipulating the features' activations. ~\cite{makelov2024principled} compares the features discovered via sparse autoencoders with those created in a supervised way on a specific task (Implicit Object Identification/ IOI), finding that SAE-discovered features are of somewhat lower quality than the supervised ones, building on~\cite{wang2022interpretability}, which reverse engineers circuits in GPT2-Small for the IOI task.

\section{Experiments with Editing Models Trained on Factual Data}

We now present a series of experimental findings from training models on synthetic sentences corresponding to $\{s, o, r\}$ tuples generated using three simple distributions: independently generated tuples, tuples with a 100\% correlation between two of the relationships, and tuples with nested relationships, as described in the subsequent sections. In all cases, we use a very simple grammar, where the subject, relationship, and object are joined using predetermined filler tokens.

Despite the simplicity of the framework, we observe several interesting patterns: that model editing is frequently \emph{more effective} when only a subset of the model layers are edited, and the optimal choice of layers depends on the editing strategy and on the size of the change; that fine-tuning the model to `forget' an entire relationship is relatively difficult without severely damaging model quality, and that interpretability techniques such as activation patching, which have been used by methods such as~\cite{meng2022locating} for model editing, have only limited utility in layer selection in our framework.

\subsection{Model selection}
For our experiments, we chose the Pythia-31m Transformer model from the GPT model family as the architecture on which to experiment. This architecture consists of six blocks of standard Q, K, V Attention and up-down MLP layers, with a hidden dimension of 256. We chose this architecture because it is large enough to generate interesting and relevant findings, but small enough that the training data size and the model training times are very manageable - in particular, we found that we can train the model from random initialization to $>95\%$ accuracy in 24 hours on an NVIDIA RTX A6000 GPU, and can then fine-tune a model to edit a fact in minutes. These fast runtimes allow us to run far more experiments than is possible on the available corpora of natural data.

\subsection{Fact generation}
The tuples are created as follows. Each subject is assigned an object at random for each of six relationships. These roughly correspond to the setup in~\cite{allenzhu2023physics}, where each subject was randomly assigned six attributes, such as a birthdate and a place of work, but unlike the setup in that paper, the data and tokens are wholly artificial and not human-interpretable. This is done to prevent token collisions. For the experiments in this paper, except where specified, we use 400 possible object values for each relationship, drawn uniformly at random for each subject and relationship. Thus, a sample tuple might look like $(125, 2, 48)$, indicating that the value of relationship 2 for subject 125 is 48. The number of subjects is chosen using simple binary search to have the model reach 95-98\% accuracy in 460 training epochs, suggesting that the model is roughly at `capacity' for the data. In the most basic setup described (6 relationships per subject with 400 possible choices of objects), the Pythia-31M architecture reaches 95\% accuracy with 120 000 subjects.

\subsection{Sentence construction}
From the tuples of facts, we construct sentences designed to loosely imitate natural language as follows. We construct sentences from one or several templates, with several additional ``grammar'' tokens taking the space of common words, and special ``SS'', ``RR'', and ``OO'' tokens directly preceding the subject, relationship, and object tokens, to aid human readability. As an intuition, if we consider a sentence like ``Speaking of the man Alexander, he currently works at Microsoft'', the additional ``grammar'' tokens would take the roles of the words ``Speaking", ``of'', and ``he''. The ``SS'', ``RR'', and ``OO'' tokens would then correspond to ``man'', ``currently'', and ``at'', and the subject, relationship, and object would be ``Alexander'', ``works'', and ``Microsoft''. A sentence in the grammar of this experiment might then look as follows: \texttt{SS 125 FT1 FT2 RR 3 FT3 OO 48.<|endoftext|>}, where \texttt{FTx} stands for the filler tokens. 

\subsection{Tokenization}

We rewrote the tokenizer from scratch to limit the expressivity of the model entirely to the tokens necessary to capture the facts and grammar we constructed. As such, each token, except for the special period and \texttt{<|endoftext|>} characters, as well as the special \texttt{SS}, \texttt{RR}, and \texttt{OO} characters, is a four-digit number preceded by a space, and an entity can consist of one or several tokens. In particular, for most experiments, the subject and object consist of two tokens, and each relationship consists of one token.

The token space is fully partitioned, with no shared tokens between the spaces of subjects, relationships, objects, and other grammar particles; further, in two-token constructions, such as the subjects, the sets of possible first and second tokens are disjoint. As a technical detail, additional tokens corresponding to all possible prefix strings of the used tokens were created, but are not used for sentence construction; the greedy tokenization always prefers the longer 4-character (or special character) tokens. The sentences corresponding to each subject-relationship-object tuple are then fully shuffled, ensuring that all information regarding each subject/relationship cannot be drawn from other relationships of that subject, a risk found in~\cite{allenzhu2023physics}.

\subsection{Model training}

We use a rough binary search to find a data size that meets our accuracy requirements after 460 epochs of training. We shuffle the data as follows. First, we create the sentences as described above. Then, we shuffle the entire set of sentences, without keeping the sentences relating to one subject together, and concatenate them into one long block of text. In all experiments, we use a fixed learning rate of $0.001$ with a batch size of 16 and a weight decay of 0.01, and an example sequence length of 512. We use the LitGPT library~\citep{litgpt-2023} for all training and fine-tuning, which we modified to be able to control which layers of the network are fine-tuned for each experiment.

\subsection{Model editing} We use fine-tuning as the principal method of model editing, as this is generally one of the most direct and practicable approaches; in addition, it allows us to redirect the undesirable output to another token, allowing us to preserve the sentence grammar. For fine-tuning, the Behemoth framework provides several options for assembling datasets consisting of the desired edits combined with a sampling of `clean' data that should remain unchanged, which we found to be essential to preserving accuracy in the rest of the model. These sentences are shuffled by the data loader during training. At the same time, the framework creates test data to verify the model's performance on other knowledge and tasks. 

We use a range of learning rates and epoch lengths for fine-tuning models, with the best learning rate and epoch length chosen individually for each experiment from all runs. For full-rank fine-tuning we train for one epoch. For low-rank fine-tuning, we tune the number of steps for each run.

\subsection{Does model editing work?}

In this section, we conduct experiments to measure the efficacy of model editing across four scenarios, which are as follows. 
\begin{itemize}
    \item First, we select one (correctly classified) $\{s, r, o\}$ triple at random, and change the value of $o$ to another possible value, also selected at random.
    \item Then, we repeat the experiment, but selecting ten triples with the same $r$ and $o$ values and replacing all those $o$ values with the same alternative $o$ value.
    \item Then, we select ten triples with different $o$ values (but the same $r$ value) and replace the objects with ten different other possible values. These three experiments investigate the ability of the model to reliably change a prediction, as well as investigate the effect on the rest of the model's predictions. 
    \item The final experiment measures whether the model can be guided to forget an entire concept, which we simulate by remapping all objects for a single relationship to the same value. We chose that definition of ``forgetting'' because it removes all information for that relationship, while staying within the rules of grammar that were created for these experiments.
\end{itemize}

We conduct these four experiments in three scenarios. We briefly introduce these scenarios here, then describe them in more detail when presenting the results of the model editing experiments.
\begin{itemize}
    \item In the first, the training data consists entirely of $\{s, r, o\}$ tuples as described above, all of which are independently generated. 
    \item In the second, we establish a link between the first and second relationship, where the value of the first entirely determines the value of the second; for simplicity, we set these two relationships to be binary (so $r_1$ and $r_2$ are set to either $(0,0)$ or $(1,1)$). The goal in this scenario is to measure whether the model editing is always comprehensive - that is, whether editing a fact or a relationship will also update a `synonym' fact, or if it is possible to recover the original information. 
    \item In the final scenario, we introduce the concept of nested relationships, where each object in our $\{s, r, o\}$ tuples is itself the subject of a different $\{o, \hat{r}, \hat{o}\}$ tuple, and therefore we can also form $\{s, \hat{r}, \hat{o}\}$ tuples, where. Ideologically, this is akin to a set of statements, akin to "Mary works at Acme Corp. Acme Corp is located in Springfield. Mary lives in Springfield". The goal in this setup, as in the one above, is to measure the direct and indirect (downstream) effects of changing $\{s, r, o\}$ tuples.
\end{itemize}

In the first three cases, the fine-tuning dataset is made up of a mixture of the following: 250 repeats of the tuple(s) to be changed (with the new value for the object), examples of other tuples with the same object value that should not be changed, and tuples drawn at random from the remaining training data in a $2:1:4$ ratio. For the fourth, 'forgetting' case, we randomly sample 5\% of the tuples with the first relationship, and add samples drawn at random from training data with other relationships in a $1:1$ ratio.

For all experiments, we do full-rank (256) and low-rank (LoRA) finetuning. For LoRA, we vary the rank from 32 to 128 and vary the learning rate and number of steps to achieve the highest model editing rate; for full-rank, we use all data and vary the learning rate. We additionally present experiments using the ROME\citep{meng2022locating} for the case of changing a single fact. This method is specifically designed for factual changes, and works by computing a change in the MLP layers of a model that edits the weights to maximally edit the activations of the edited fact while minimizing the change effect elsewhere.

\subsection{Simple dataset}

We first experiment on a dataset trained purely on sentences as described above. Following the work of~\cite{allenzhu2023physics}, we have six relationships for each subject, and for each relationship, we select randomly from 400 possible values for each object. We then vary the number of subjects to create a model that is close to saturated, with accuracy over 95\% for predicting the correct object for a subject-relationship pairing. We found that using $120,000$ subjects results in 95.38\% accuracy when trained for 460 epochs (each datum is seen 460 times). We note that this does not saturate accuracy, although it comes close: training for 1000 epochs results in an accuracy of 98\%.

We briefly estimate the size of the information that is contained in the model. Neglecting the (very simple) grammar, we note that 120 000 subject values can be stored in 17 bits, six relationships can be stored in 3 bits, and 400 objects can be stored in nine bits - thus, the information could be stored in 29 bits $\times$ 120 000 subjects $\times$ 6 relationships = approximately 21 million bits, giving us a ratio of $0.65$ bits per parameter, substantially lower than $3.6$ bits per parameter estimated by~\citep{Morris2025HowMD}. We believe that this is due to the model inefficiently using its parameter space to store the highly symmetric data, and we provide evidence for this in Appendix \ref{sec:all_activation_patching}. 

We now conduct the four experiments described above: updating a single datapoint; updating ten datapoints from and to the same object; updating ten datapoints from and to different objects; and `forgetting' a relationship by setting all objects for that relationship to the same value. For simplicity, we always experiment on the first relationship, and for the `forgetting' experiment, we set all subjects to the first object possible for that relationship.

For the case of a single override, we also explore the ROME method of~\cite{meng2022locating}, which, the authors show, performs on par with the best low-rank finetuning methods for both remapping success and remaining accuracy in real-world LLMs.

\begin{figure}
    \centering
    \includegraphics[width=0.45\linewidth]{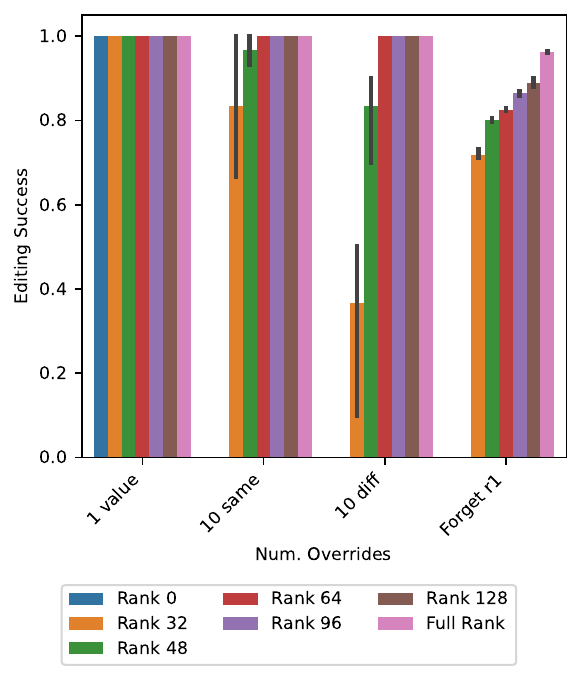}
    \includegraphics[width=0.45\linewidth]{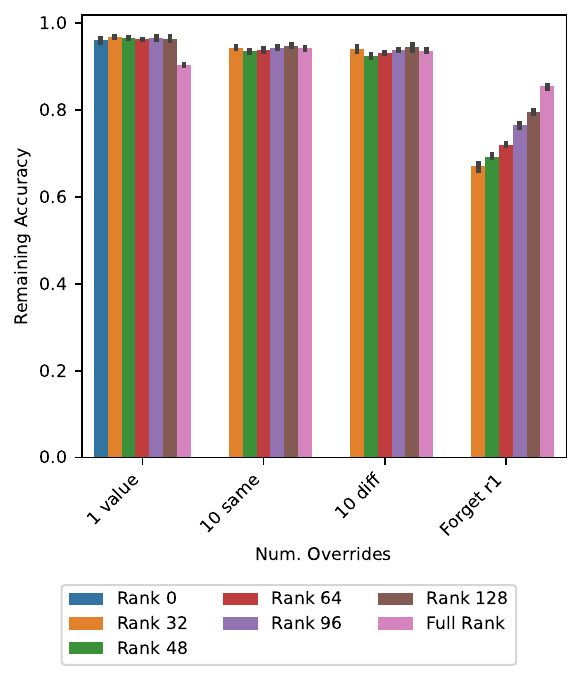}
    \caption{Success of ROME, full, and LoRA finetuning for the simple dataset scenario. All results are averaged across three runs.}
    \label{fig:simpleds_effectiveness}
\end{figure}

The results of the experiments are shown in Figure~\ref{fig:simpleds_effectiveness}. Here and below, we define ``Editing success'' to denote that the model now outputs the desired value as the top-probability token (or the first two top-probability tokens) when prompted with the beginning of the phrase. We observe that when only a single model edit or ten identical edits are made, it is already possible to make the edit and retain high remaining data accuracy with a rank-32 update, although editing with ROME fails about 5\% of the time. However, when making ten \emph{different} edits, we need at least a rank of 64 to successfully complete the edits; further, increasing the rank of the update decreases the remaining accuracy of the model. For the task of `Forgetting' the first relationship, i.e., remapping all objects for that relationship to a single value, we observe that using 5\% of the data points as the fine-tuning data is generally sufficient to effect the change across nearly all $\{s, r_1, o \}$ tuples, and so the fine-tuning generalizes from the sample in the data. However, the accuracy of the rest of the model is severely impacted, with resulting model accuracy ranging from 70-80\% depending on the rank of the update. For this task, rank 128 and full-rank updates were the most effective at making the update while preserving model accuracy. 

Note that this is a fairly surprising result, as, in general, the model could have achieved the desired output simply by keeping its internal representation of the data intact, and learning to output the tokens corresponding to $o_1$ for any query for $r_1$, which might have been considered a very lightweight change. Instead, it took considerable training data and was only possible with substantial changes to the model, as shown by the lowered accuracy for other relationships.

\subsection{Correlated relationships}

We then create a new version of the training data with a perfect correlation between the first and second relationships. Precisely, we binarize both relationships, so each subject can only have two possible objects for $r_1$ and $r_2$, and 400 for $r_3-r_6$. We then establish a perfect correlation between the values for $r_1$ and $r_2$, so that the values for those two relationships are either $(0, 0)$ or $(1, 1)$. For this experiment, we found that using $170 000$ subjects resulted in a model with $98\%$ accuracy after 460 training epochs. Using the same calculation as above, we derive a ratio of 0.72 bits per parameter for this setup.

We then repeat the three value changing experiments and the relationship forgetting experiment. Specifically, we only try to change the object value of $r_1$, creating training data as before. We then measure the impact of the change on $r_1$ and $r_2$ for the affected tuples, and on the remaining (`clean') data to estimate model impact.

\begin{figure}
    \centering
    \includegraphics[width=0.32\linewidth]{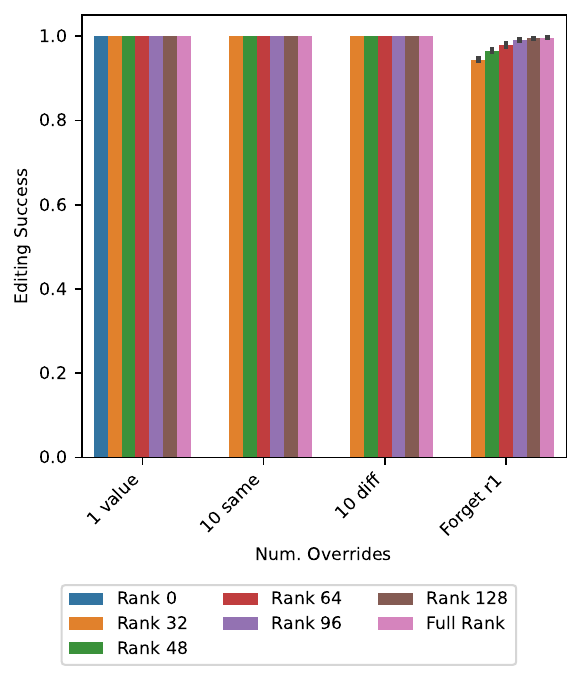}
    \includegraphics[width=0.32\linewidth]{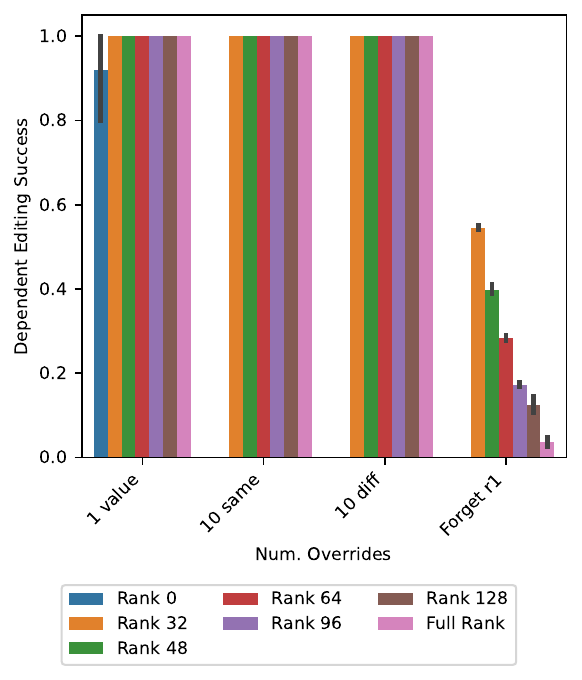}
    \includegraphics[width=0.32\linewidth]{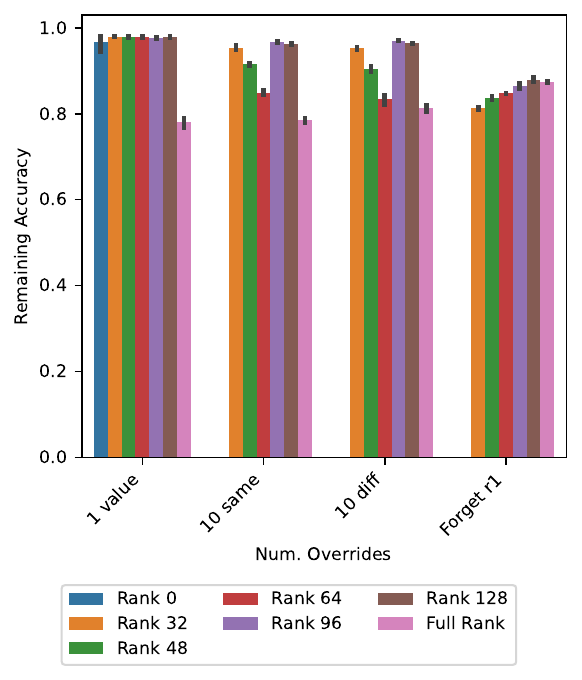}
    \caption{Success of ROME, full, and LoRA finetuning for the correlated relationship dataset scenario. All results are averaged across three runs.}
    \label{fig:correlatedds_effectiveness}
\end{figure}

The results of the experiments are shown in Figure~\ref{fig:correlatedds_effectiveness}. We observe a similar degree of success for all types of changes to $r_1$, and, as before, ROME performs well compared to high-rank and full finetuning. However, we observe that even in the case of two perfectly correlated relationships, remapping the label of $r_1$ does not necessarily remap the label of $r_2$. This is particularly apparent with ROME, where remapping a single tuple $\{s, r_1, o_1\}$ was effective 100\% of the time, but the fully dependent $\{s, r_2, o_2\}$ tuple was only updated 90\% of the time. Further, for the relationship forgetting experiment, while the effectiveness of forgetting $r_1$ is generally high (though increasing with rank) for low-rank and full finetuning, we observe that the `forgetting' of the fully dependent ${s, r_2, o_2}$ tuple \emph{decreases} with finetuning rank, and is almost completely ineffective when done with full finetuning, suggesting that in this scenario the update resulted in a network weight change that edited the model output without removing the original information from the model. (Recall that the `forgetting' task is trained only on $r_1$ tuples and not $r_2$ tuples). From this, we conclude that while the neural network does, to a large extent, learn the fully dependent relationship between $r_1$ and $r_2$, the storage of this information is not perfect, and the two can be decoupled during finetuning. Further, this agrees with other works~\citep{Wu2024EvaluatingDU, Joshi2024TowardsRE, Hong2024DissectingFU} that show that fine-tuning for forgetting is brittle and can reveal other relationships - our novel contribution here is to demonstrate that this happens even in cases of perfect information dependency, as in this toy scenario, and can depend heavily on finetuning hyperparameters in unexpected ways.

\subsection{Nested relationships}

Finally, we repeat the experiment with nested relationships. Specifically, as before, we assign to each subject $s$ and relationship $r$ an object $o$ drawn from a set of 400. Additionally, for each object $o$, we create a relationship $\hat{r}_o$, and randomly select an object $\hat{o}$ from a new set of 40. This sets up an implied relationship $\hat{r}_s$ between $s$ and $\hat{o}$. We train the model on a mix of all three tuples - $\{s, r, o\}$, $\{s, \hat{r}_s, \hat{o}\}$, and $\{o, \hat{r}_o, \hat{o}\}$. (We use the `hat' notation to denote nested relationships and objects from here on.) Intuitively, this is meant to model a body of knowledge with statements like "Mary works at Acme Corp", "Acme is located in Springfield", "Mary lives in Springfield".

In this case, we find that it is only possible to train an accurate (97\%) model of around $80000$ subjects, a $1/3$ drop from the `simple' model. This suggests that the implied relationship between objects and nested objects is not fully captured (or `grokked'): in principle, the bit-cost of storing the object-nested object relationship is very low: $(log_2(400) + log_2(6) + log_2(40)) \times 400 \times 6 = 41 000$, and from there, the implied subject-nested object relationship could have been inferred. We see more evidence of this in the model editing experiments.

To ensure that the inability to `grok' the data structure is not due to insufficient training, we attempted to train a model on 120 000 tuples for five times longer, i.e., showing each example to the model 2 300 times. Even in this case, the model only reached 85\% accuracy for Subject-Object mappings and 88\% accuracy for Subject-Nested Object mappings, showing that even training for a very long time does not allow the model to store this nested data efficiently. We leave it to future work to establish if providing many additional examples (i.e., drawing from an unlimited distribution of subjects with related objects) would enable the dependent relationship to be `grokked'.

As before, we perform the experiments described above. Specifically, we override one or ten $\{s, r, o\}$ tuples as before and investigate the effects of the override. For the `forgetting' experiment, we unlearn $r_1$, the first relationship, as before.

\begin{figure}
    \centering
    
    \includegraphics[width=0.32\linewidth]{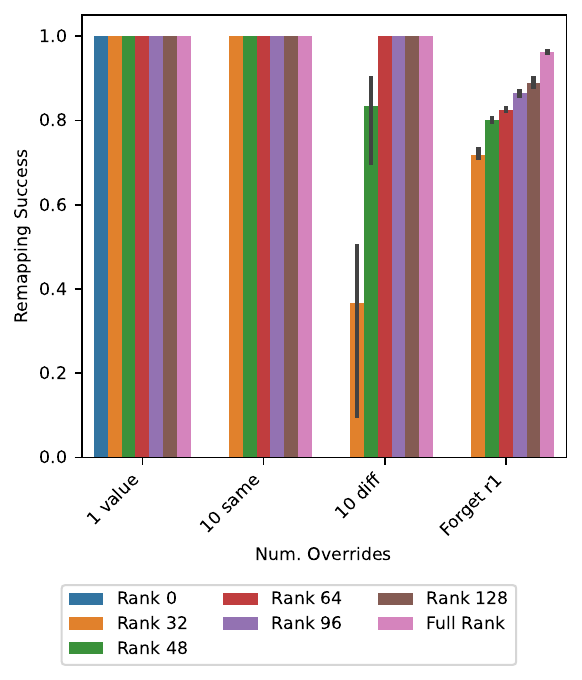}
    \includegraphics[width=0.32\linewidth]{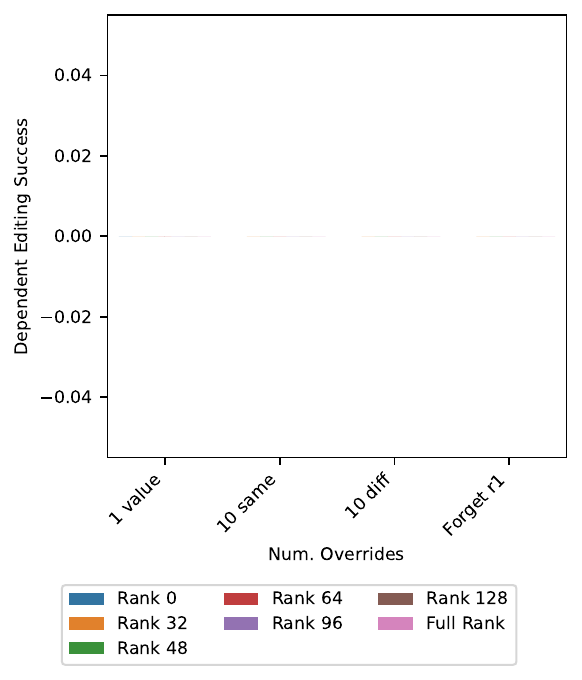}
    \includegraphics[width=0.32\linewidth]{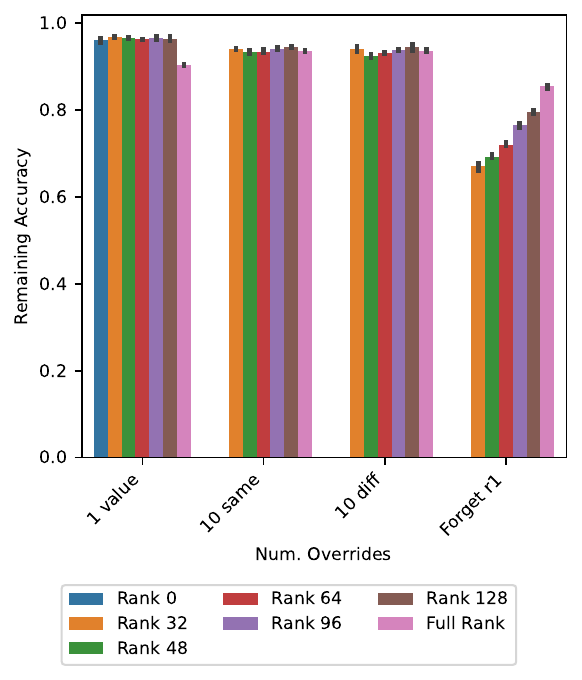}
    \caption{Success of ROME, full, and LoRA finetuning for the nested relationship dataset scenario. All results are averaged across three runs.}
    \label{fig:nestedds_effectiveness}
\end{figure}

The results of these experiments are presented in Figure~\ref{fig:nestedds_effectiveness}. We observe that, as before, creating single or ten overrides is completely effective for the single and '10 same' override scenarios. When ten different overrides are made, the update must be rank 64 or higher to achieve the same effectiveness with the same step count. Like in the simple scenario, near-perfect forgetting of $r_1$ is only possible with a full-rank update, although even a rank 32 update achieves 70\% success at this task. However, while the accuracy of the remaining $\{s, r_1, o_1\}$ tuples remains high for the tuple editing task, accuracy drops for the relationship forgetting task, with the lower-rank updates resulting in a higher forgetting rate.

We further note that, in all cases, the editing success of the dependent relationship (i.e., the tuple $\{s, \hat{r}_1, \hat{o}_1\}$) is practically 0. In particular, forgetting ${r_1}$ has no effect on forgetting ${\hat{r}_1}$.

\section{Is it necessary to fine-tune the whole model?}
\label{sec:sub_blocks}

The choice of using a 31B-parameter model that takes several minutes to fine-tune allows us to conduct many experiments in order to better understand what happens during model training and editing. Therefore, we experiment with only fine-tuning some of the layers. In particular, we experiment with training all subsets of the six transformer blocks, and with training only MLP layers, only Attention layers, or both. 

\subsection{Simple dataset}

\begin{figure}
    \centering
    \includegraphics[width=0.2\linewidth]{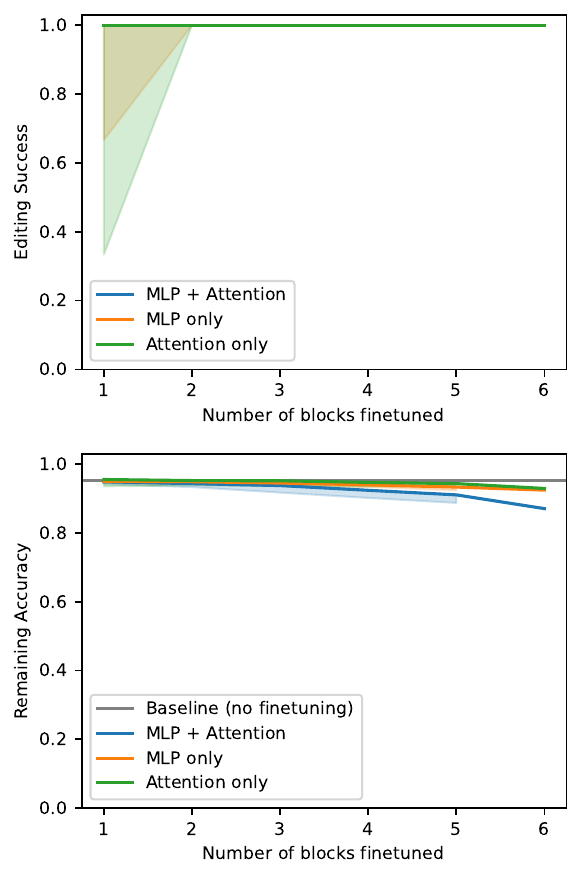}
    \includegraphics[width=0.2\linewidth]{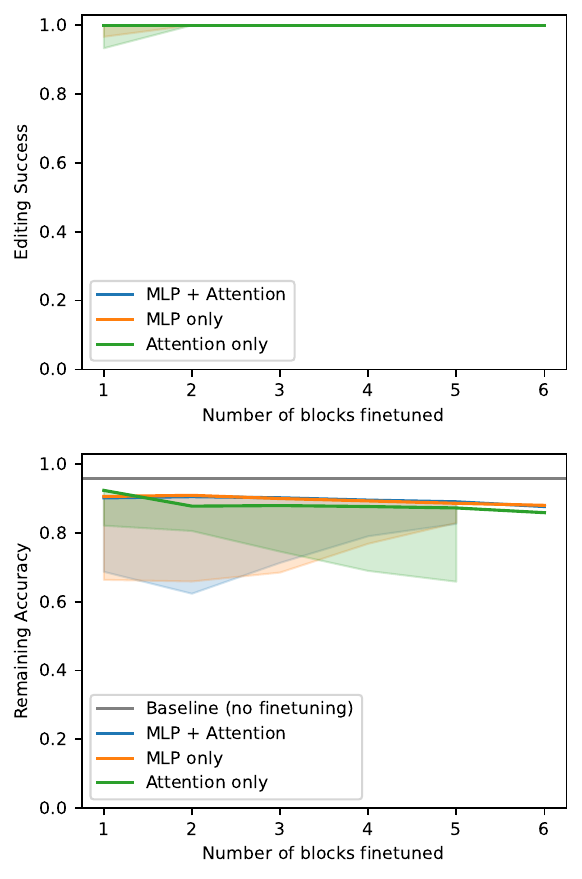}
    \includegraphics[width=0.2\linewidth]{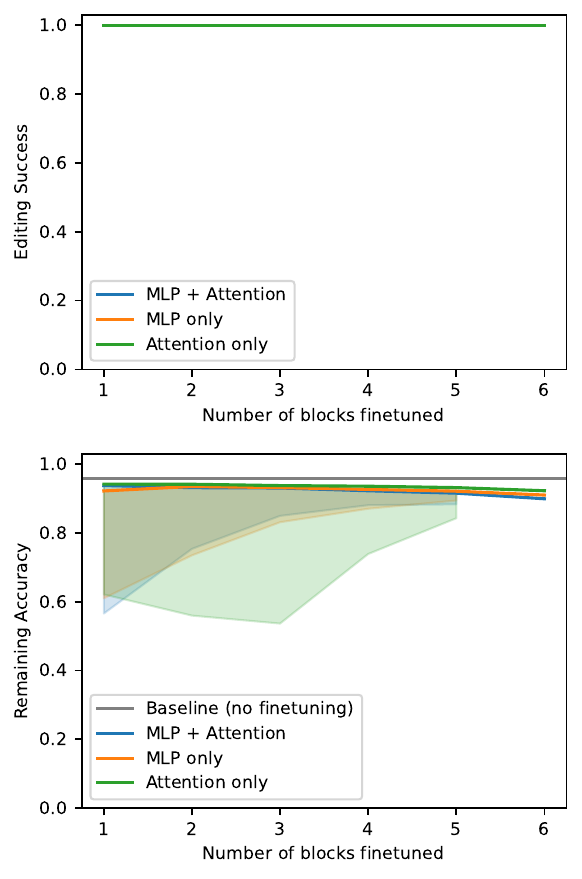}
    \includegraphics[width=0.2\linewidth]{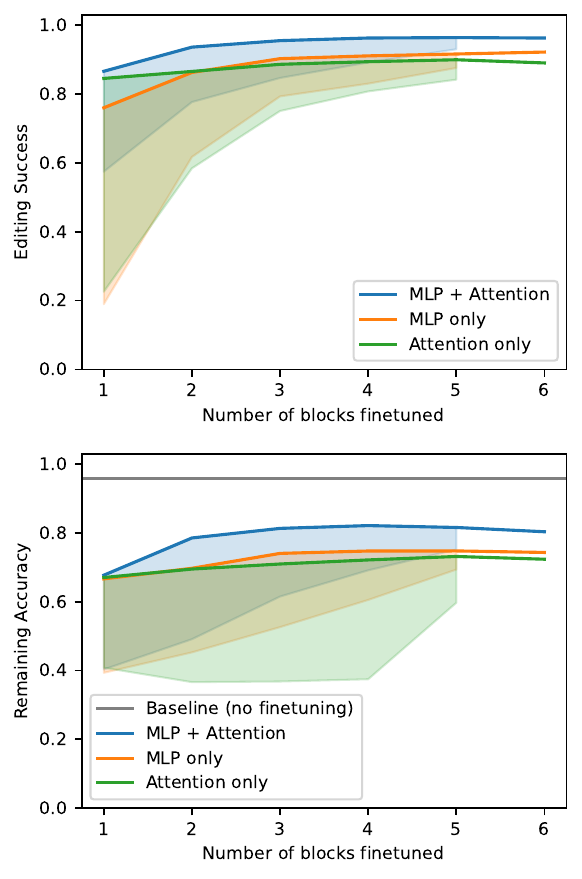}
    \caption{Simple dataset. Ability to effect the change (top) while preserving the rest of the model accuracy (bottom) of, from left to right, making a single override, ten of the same overrides, ten different overrides, and forgetting a relationship.}
    \label{fig:simple_dataset_impact}
\end{figure}

We present the results of full fine-tuning of various subsets of layers in Figure~\ref{fig:simple_dataset_impact}. In this graph, we show the effect of fine-tuning a varying number of blocks, from one to six, and fine-tuning just the MLP layers, just the Attention layers, or both. The solid line shows the best result for that number of layers, while the shaded region covers the full spread of results across all block subsets of that size. We observe that for all one- or ten-tuple editing tasks, fine-tuning just the MLP or Attention layer of a single block is sufficient to achieve the change while maintaining accuracy on the rest of the model. Further, any combination of two or more blocks is sufficient to fully achieve the desired edit, even if only the MLP layers or only the attention layers of those blocks are edited. However, for the tasks where ten tuples are edited, the choice of blocks is more important: specifically, editing the first or last MLP layer, or the last attention layer, results in a run with the highest remaining accuracy. Note also that when fine-tuning only the MLP or only the Attention layers of the model, the best-performing models (highest remaining accuracy at 100\% editing success) occur when fine-tuning only part of the model: only the first and last block when editing only MLP layers, and only the last block when editing attention layers.

\begin{figure}
    \centering
    \includegraphics[width=0.24\linewidth]{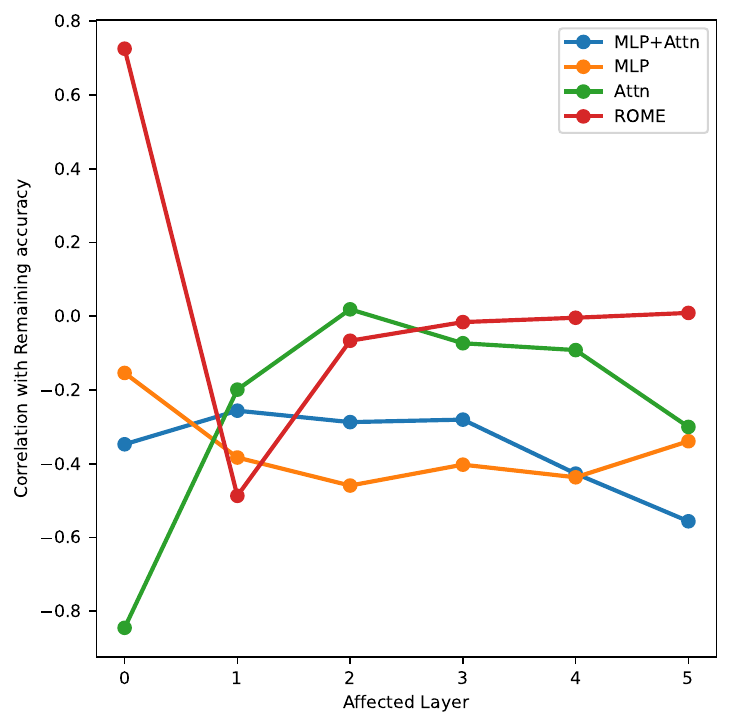}
    \hspace{1em}
    \includegraphics[width=0.24\linewidth]{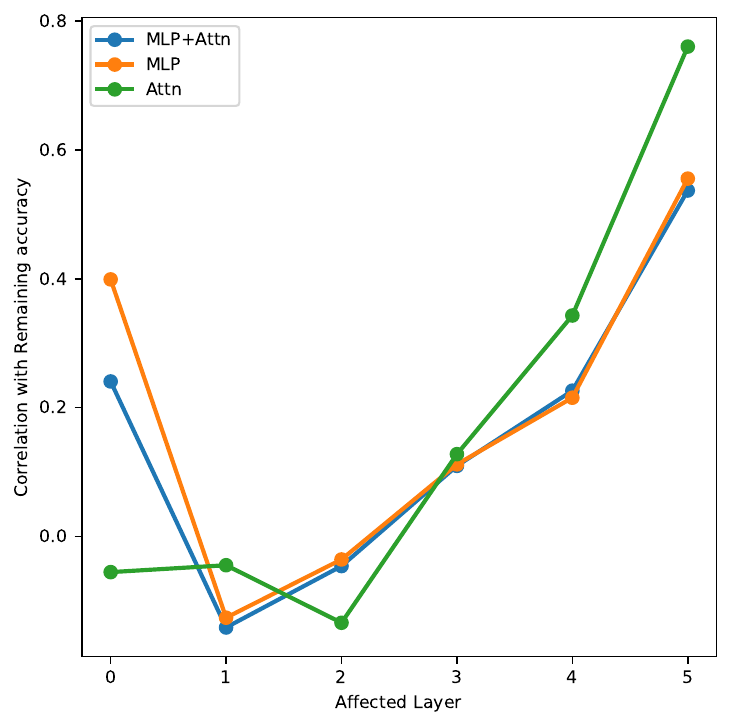}
    \hspace{1em}
    \includegraphics[width=0.24\linewidth]{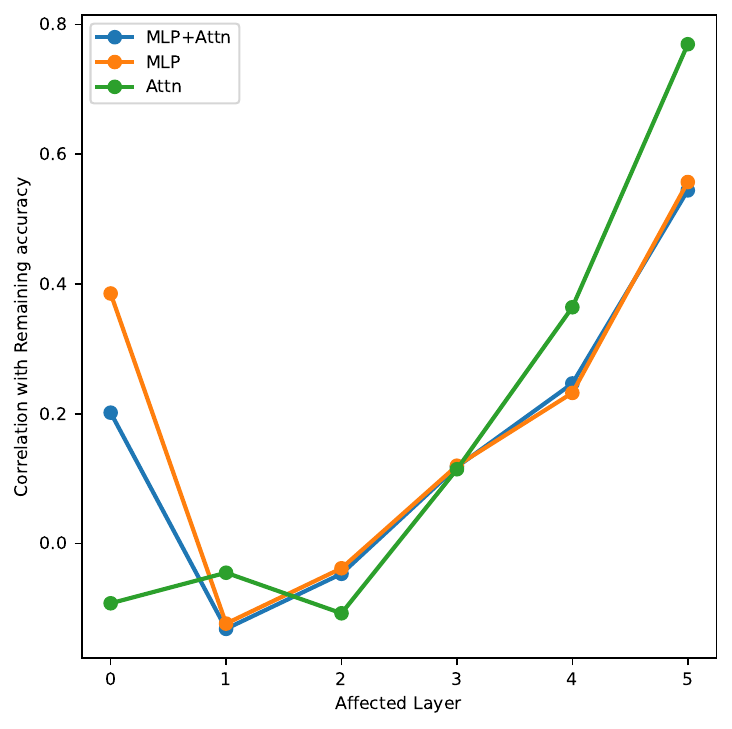}
    \caption{Simple dataset. Correlation of whether a block is fine-tuned with remaining accuracy, for editing a single tuple (left), editing ten tuples the same way (middle), and making ten different edits on ten different tuples (right).}
    \label{fig:simple_dataset_correlations_editing}
\end{figure}

We then examine the marginal impact of training each block by measuring the correlation between editing this block and the overall success of the model edit. In Figure~\ref{fig:simple_dataset_correlations_editing}, we observe that, when editing a single tuple, all blocks have a negative correlation with the resulting model accuracy. This is especially true for the last two blocks (when editing MLP and attention), the last 5 blocks (when editing MLP layers only), and the first block (when editing attention only). In general, for this scenario, the fewer blocks are fine-tuned, the better the resulting model. When the editing complexity rises to editing ten tuples, however, the situation reverses. Correlations for fine-tuning blocks generally become positive, with the last block and the MLP of the first block especially influential in preserving model quality. Note that this occurs despite the fact that \textit{the dataset creation process is the same in all cases - that is, the fine-tuning dataset for editing ten tuples is simply ten datasets for editing a single tuple, concatenated and shuffled.}

\begin{figure}
    \centering
    \includegraphics[width=0.24\linewidth]{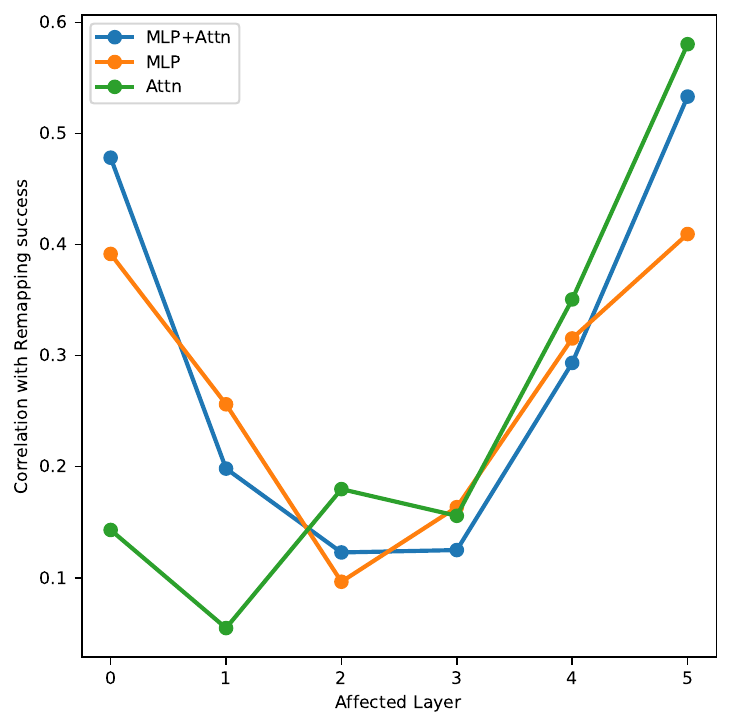}
    \includegraphics[width=0.24\linewidth]{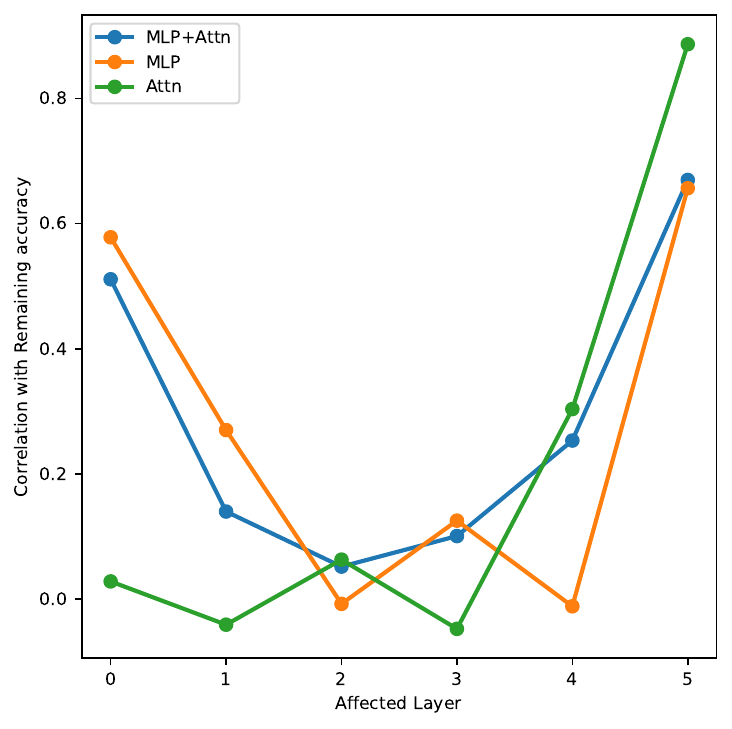}
    \caption{Simple dataset. Correlation of whether a block is fine-tuned with editing success (left) and remaining accuracy (right), when forgetting an entire relationship.}
    \label{fig:simple_dataset_correlations_forgetting}
\end{figure}

\begin{figure}[h]
    \centering
    \includegraphics[width=0.4\linewidth]{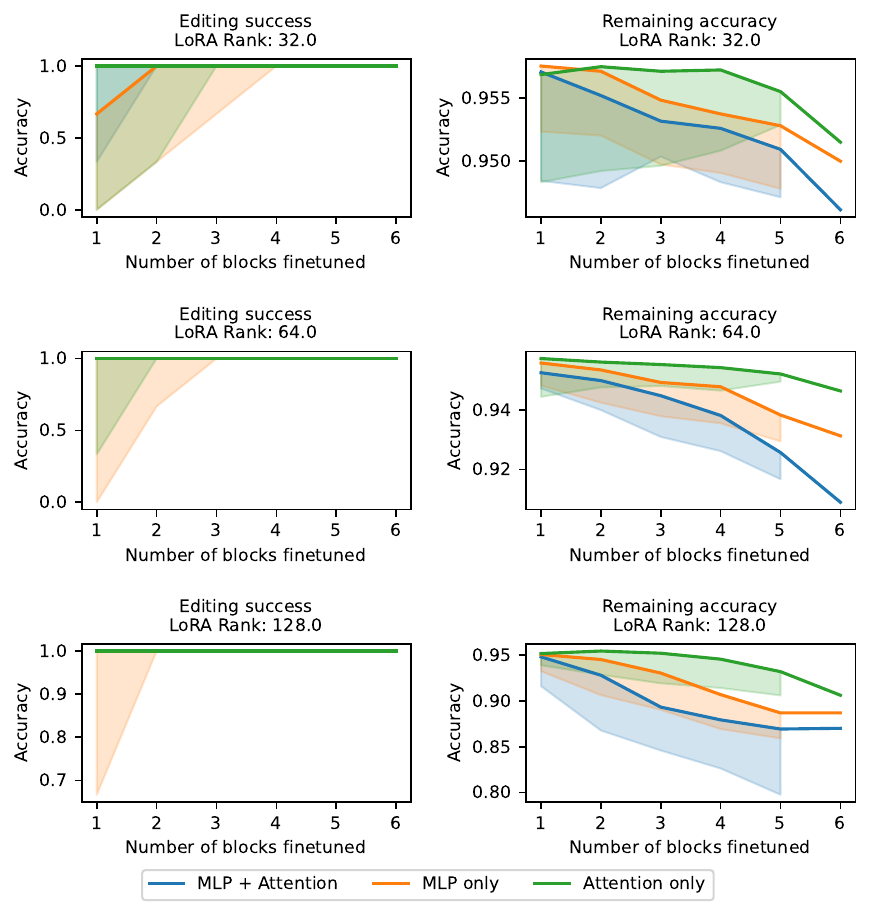}
    \hspace{1em}
    \includegraphics[width=0.4\linewidth]{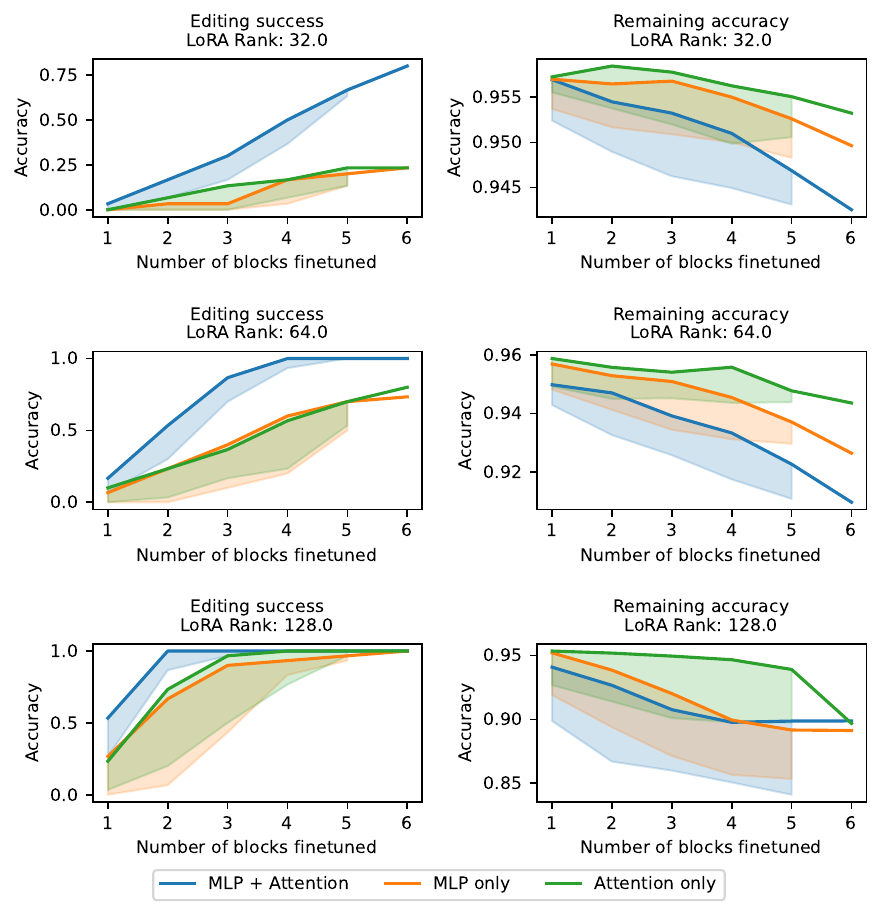} \\
    \vspace{1em}
    \includegraphics[width=0.4\linewidth]{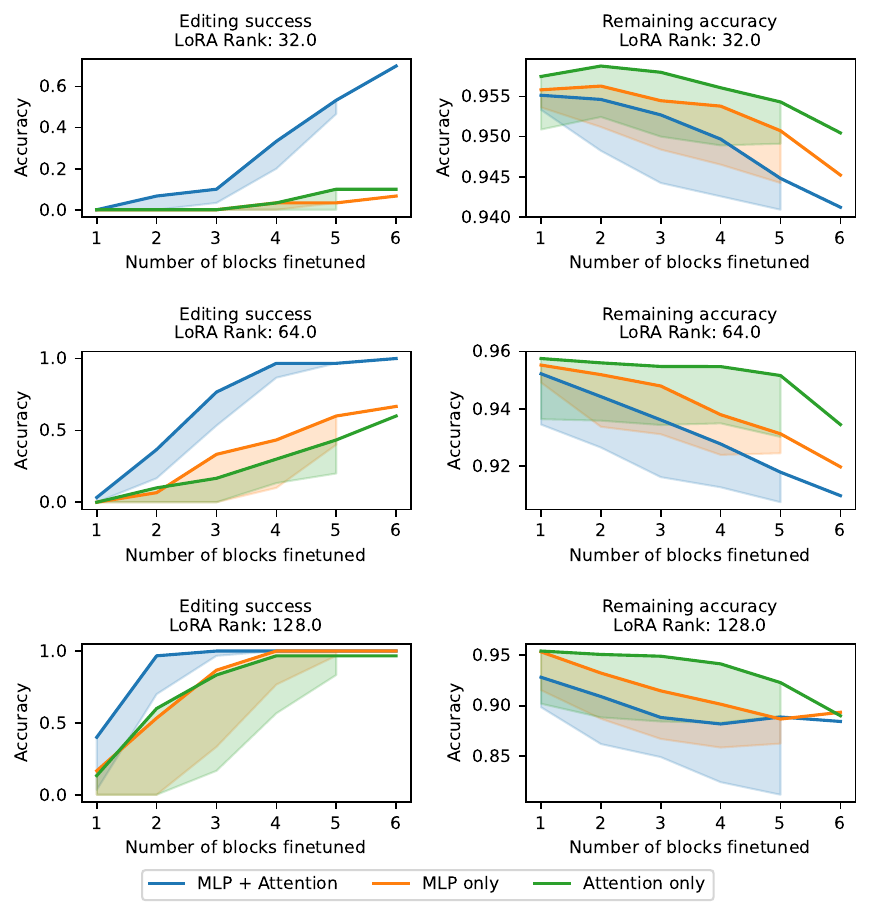}
    \hspace{1em}
    \includegraphics[width=0.4\linewidth]{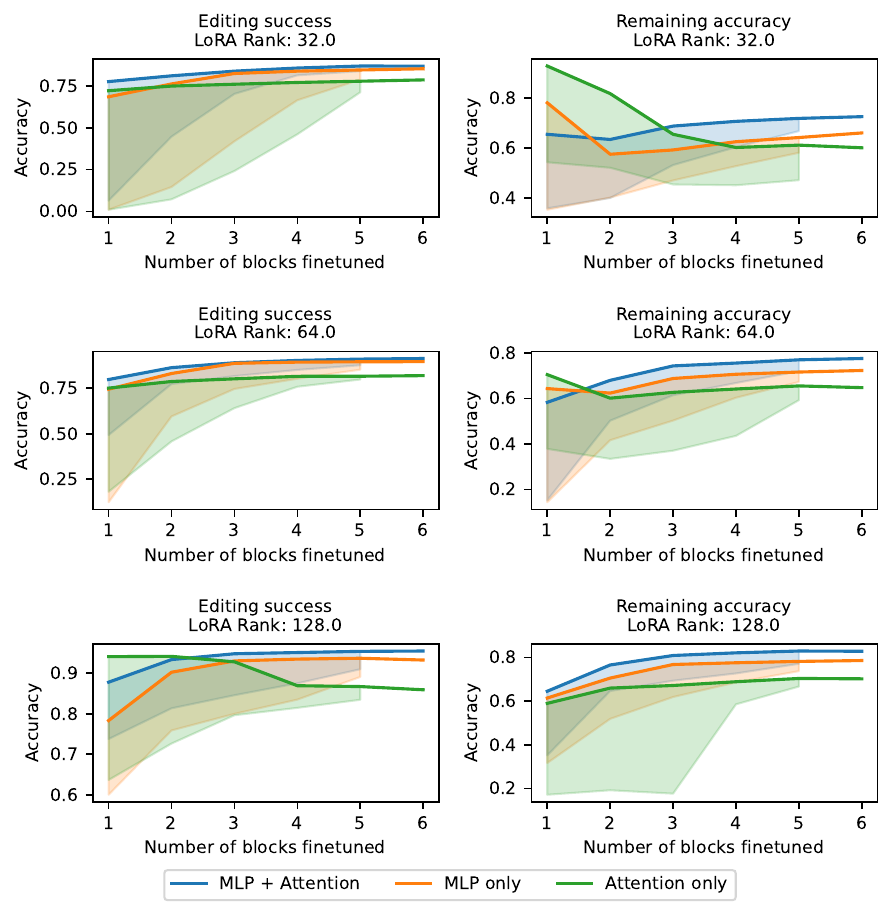}
    \caption{Simple dataset. Results of editing a single tuple (top, left), making ten identical edits to ten tuples (top, right), making ten different edits, and forgetting a relationship with LoRA at various ranks, for 400/400/400/2500 training steps. }
    \label{fig:single_lora}
\end{figure}

For LoRA finetuning, shown in Figure~\ref{fig:single_lora}, we observe that, for the simpler tasks of making one or several edits, it suffices to only fine-tune several of the blocks, even at low rank. We further note that for this dataset, fine-tuning only attention layers tends to preserve more accuracy than fine-tuning MLPs or all layers, even at the same edit success rate. This is not true for the task of forgetting an entire relationship, however, where the best success-remaining accuracy tradeoffs are achieved by fine-tuning all layers or only the MLP layers. 

\subsection{Correlated relationships}
\label{sec:corr_dataset_layer_impact}

\begin{figure}
    \centering
    \includegraphics[width=0.2\linewidth]{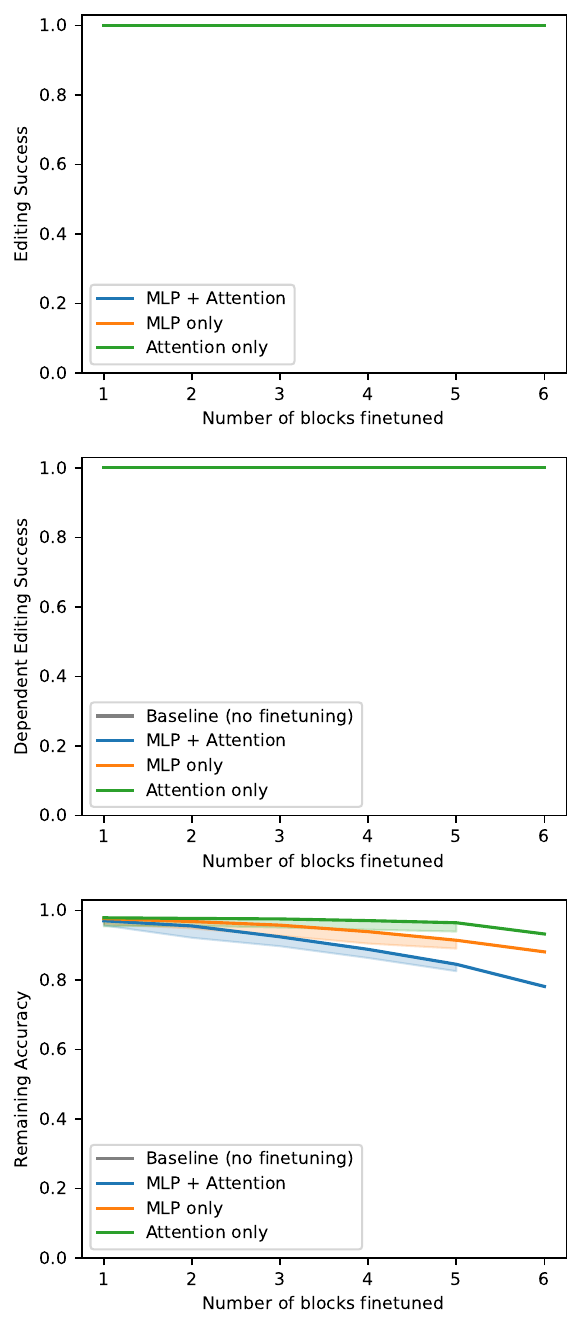}
    \hspace{1em}
    \includegraphics[width=0.2\linewidth]{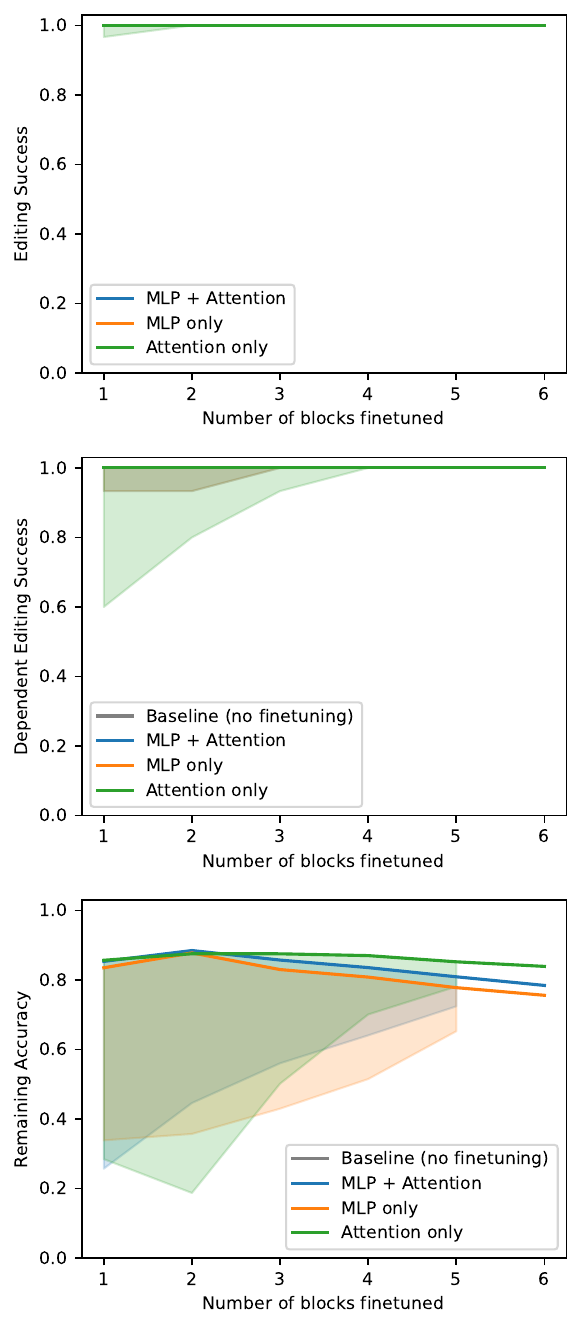}
    \hspace{1em}
    \includegraphics[width=0.2\linewidth]{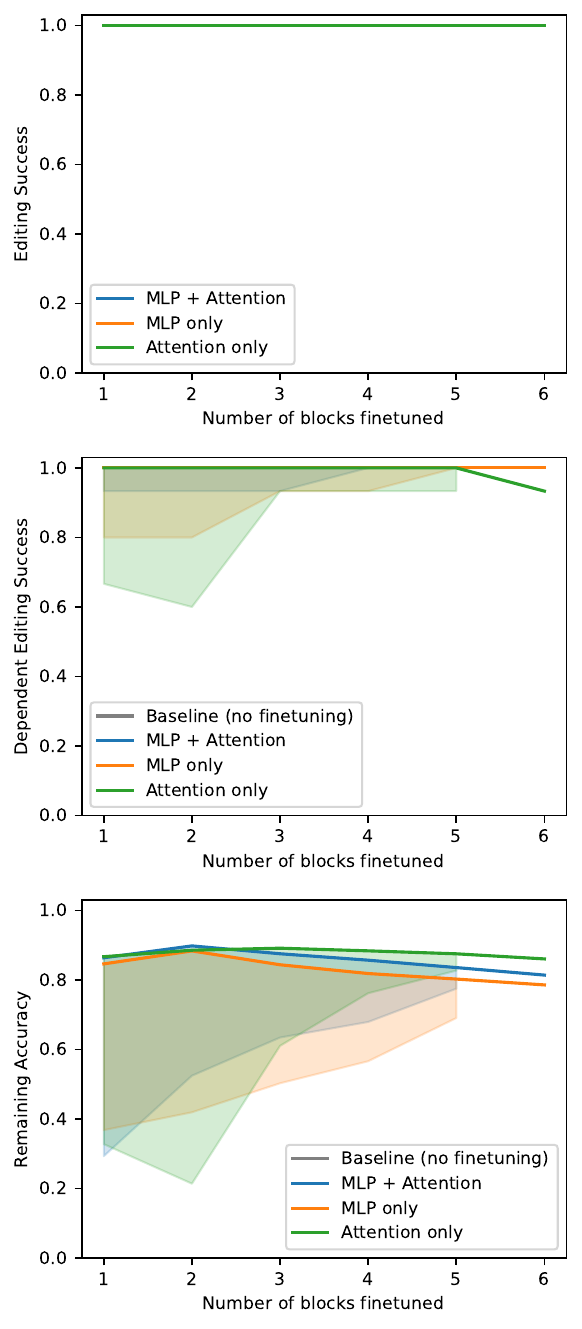}
    \hspace{1em}
    \includegraphics[width=0.2\linewidth]{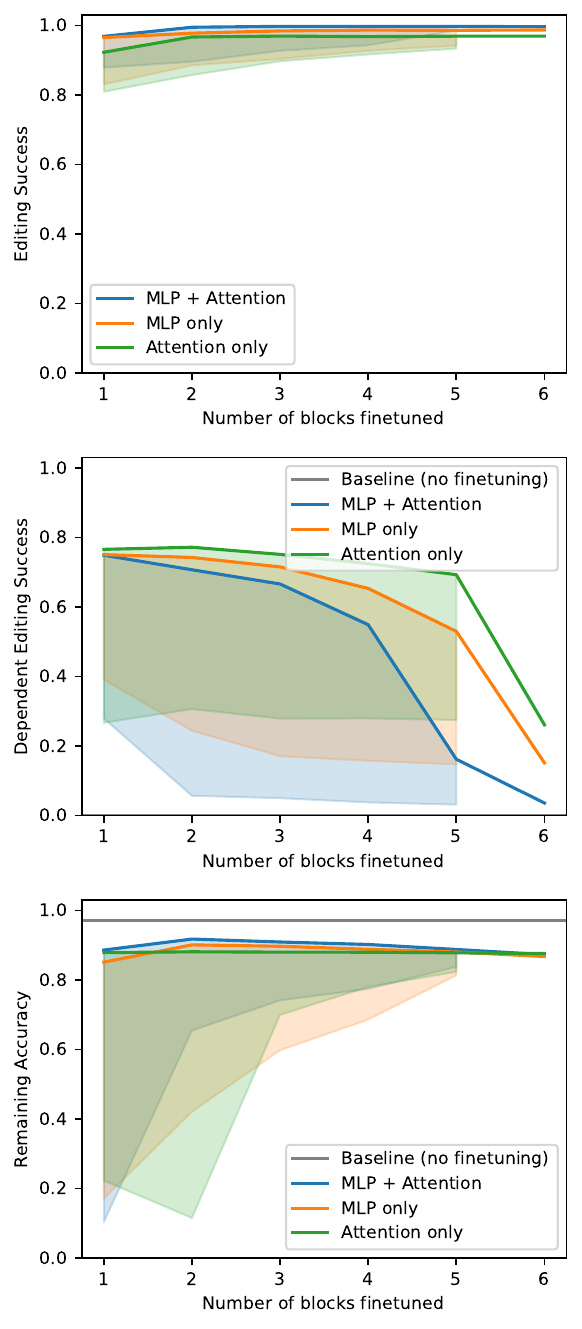}
    \caption{Correlated dataset. Ability to effect the change (top), effect the change on the dependent relationship (middle), while preserving the rest of the model accuracy (bottom) of, from left to right, making a single override, ten of the same overrides, ten different overrides, and forgetting a relationship.}
    \label{fig:correlated_dataset_impact}
\end{figure}

When we consider the correlated dataset (Figure~\ref{fig:correlated_dataset_impact}), we observe that, while it is possible to edit one or multiple tuples by fine-tuning only the MLP or only the attention layers of a single block - and, for a single edit, by fine-tuning only the MLP or only the attention layers of any block, the choice of block makes it possible to inadvertently decouple the fully correlated first and second relationship when multiple tuples are edited in the same way. For forgetting an entire relationship, it becomes crucial to fine-tune both the MLP and attention layers of the network, although it is still not necessary to fine-tune every block of the model.

\begin{figure}
    \centering
        \includegraphics[width=0.2\linewidth]{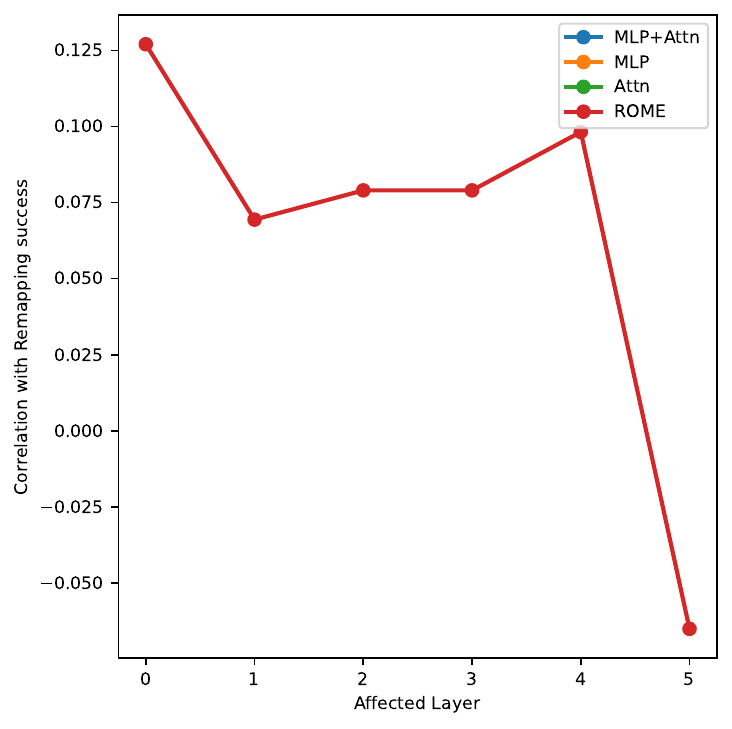}
    \hspace{0.4\linewidth}
    \hspace{2.6em}
    \hspace{1em}
    \includegraphics[width=0.2\linewidth]{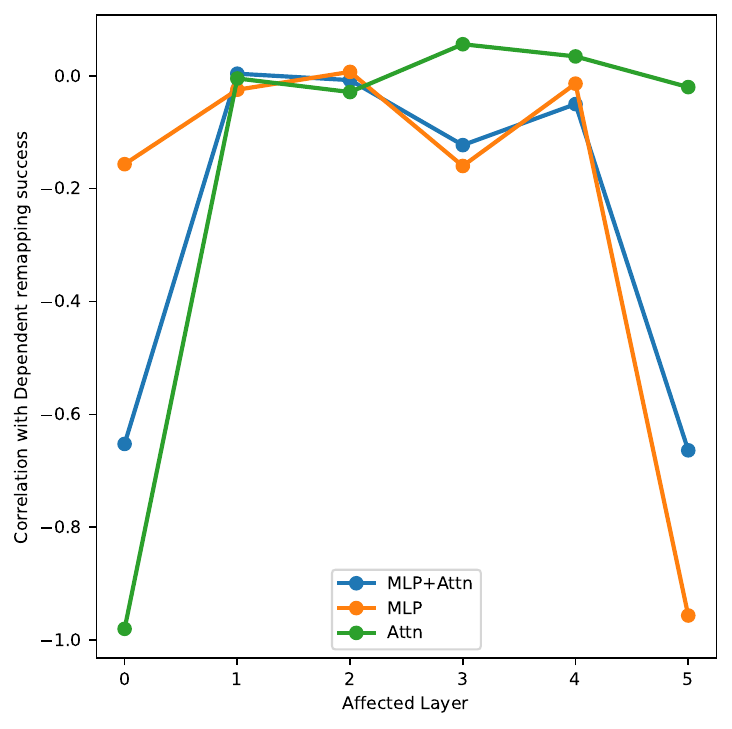}
    \\
    \includegraphics[width=0.2\linewidth]{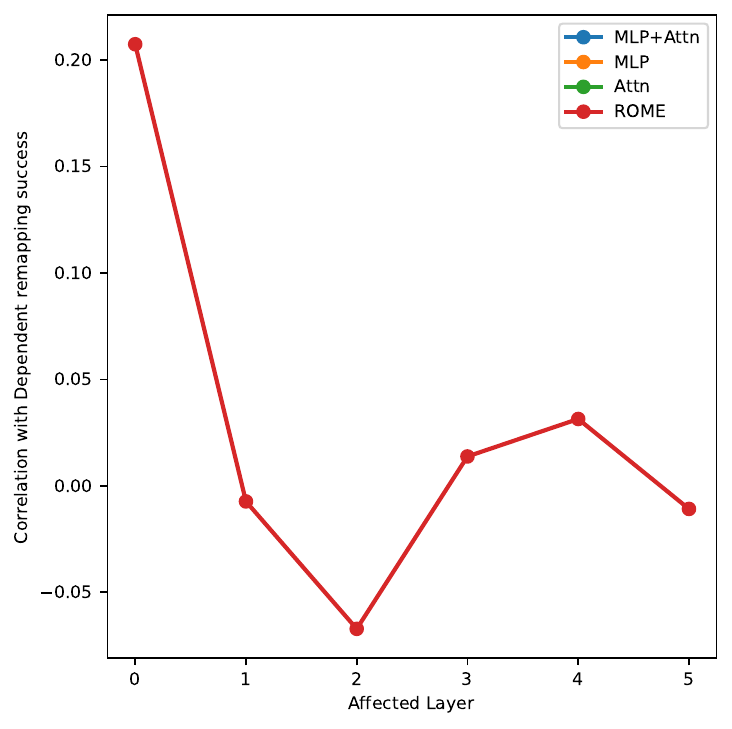}
    \hspace{1em}
    \includegraphics[width=0.2\linewidth]{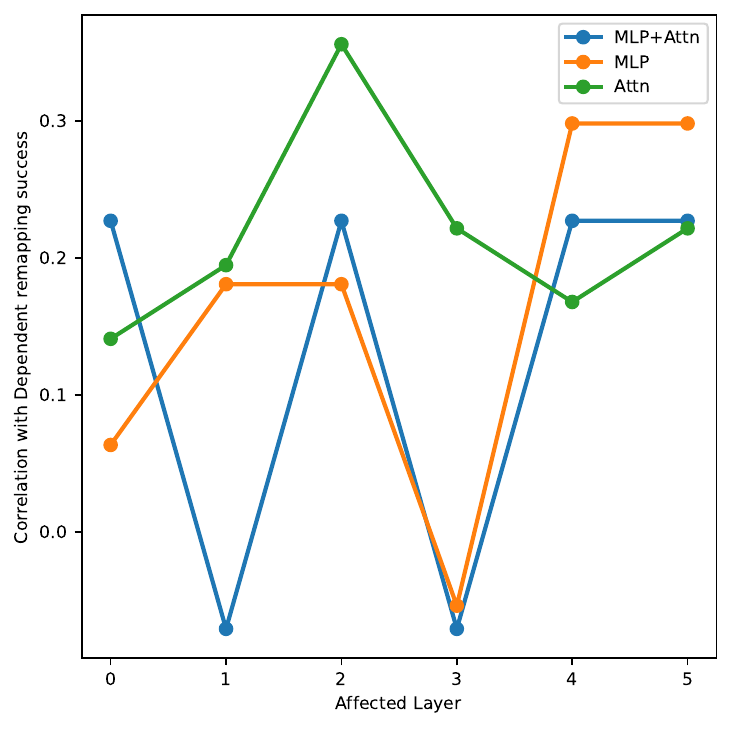}
    \hspace{1em}
    \includegraphics[width=0.2\linewidth]{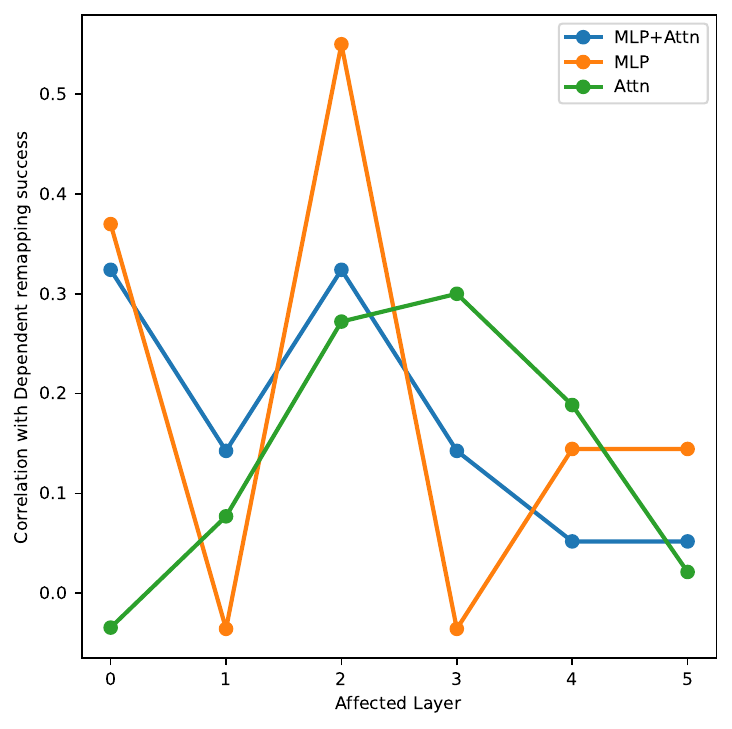}
    \hspace{1em}
    \includegraphics[width=0.2\linewidth]{graphics/performance_plots/tokenized_correlated_pair_binarized_corrstrength100_shuffled_s170000_r6_o400_n500_i0_m0/fft_ftremapping_single_subset0.05_wregdata_Dependent_Remapping_Success_layerimpact.pdf}
    \\
        \includegraphics[width=0.2\linewidth]{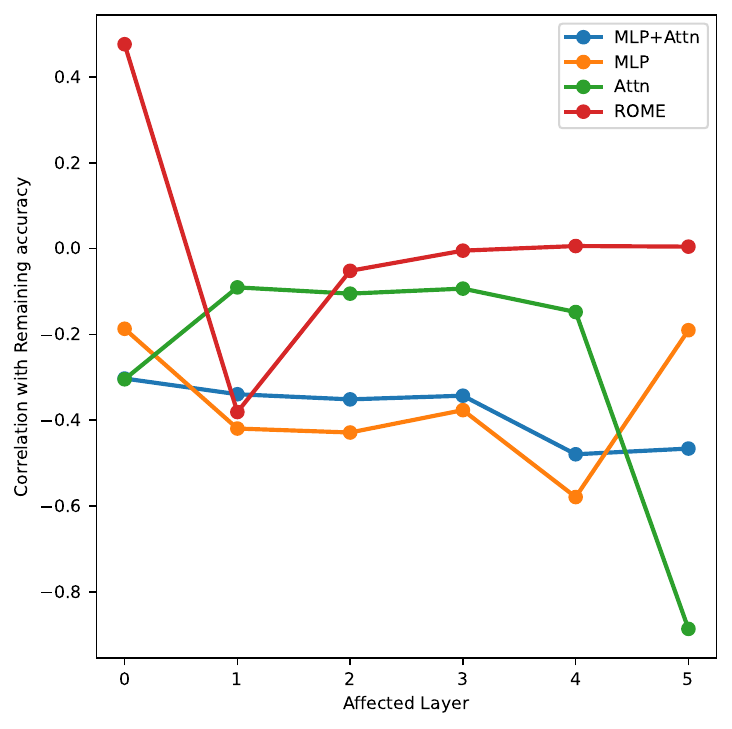}
    \hspace{1em}
    \includegraphics[width=0.2\linewidth]{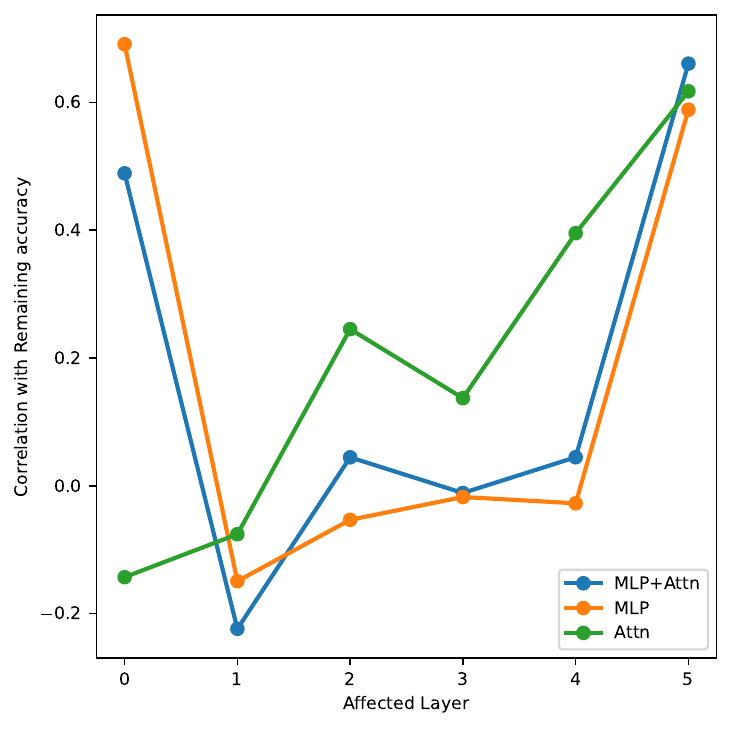}
    \hspace{1em}
    \includegraphics[width=0.2\linewidth]{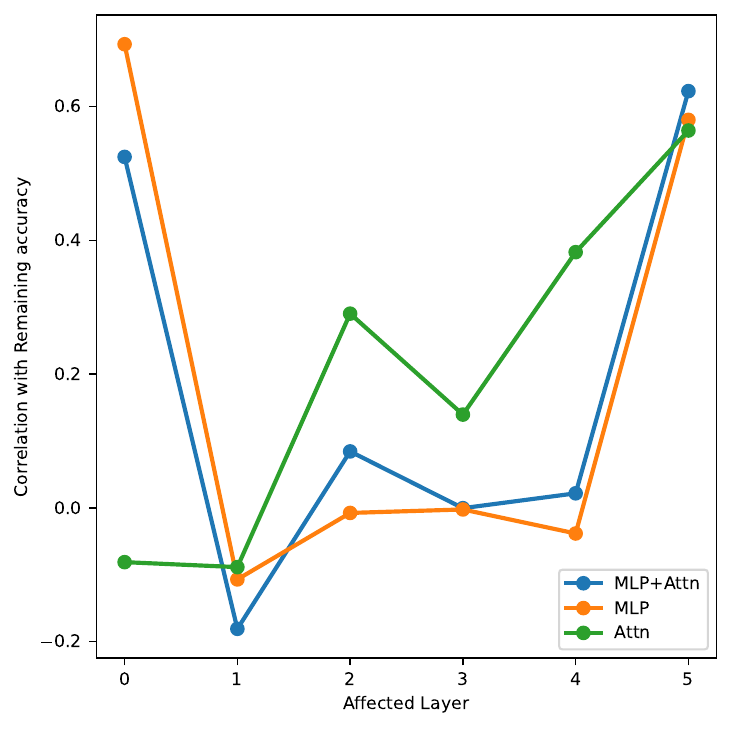}
    \hspace{1em}
    \includegraphics[width=0.2\linewidth]{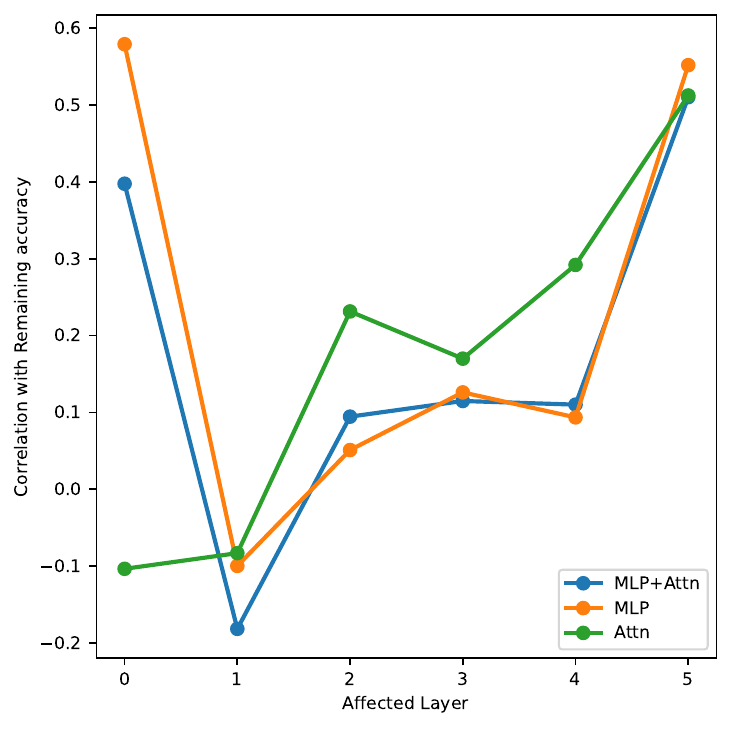}
    
    \caption{Correlated relationships dataset. Correlation of whether a block is fine-tuned with remaining accuracy, for editing a single tuple (left), editing ten tuples the same way (middle), and making ten different edits on ten different tuples (right), for dependent remapping success (top) and remaining data accuracy (bottom).}
    \label{fig:correlated_dataset_correlations_editing}
\end{figure}

\begin{figure}[h]
    \centering
    \includegraphics[width=0.45\linewidth]{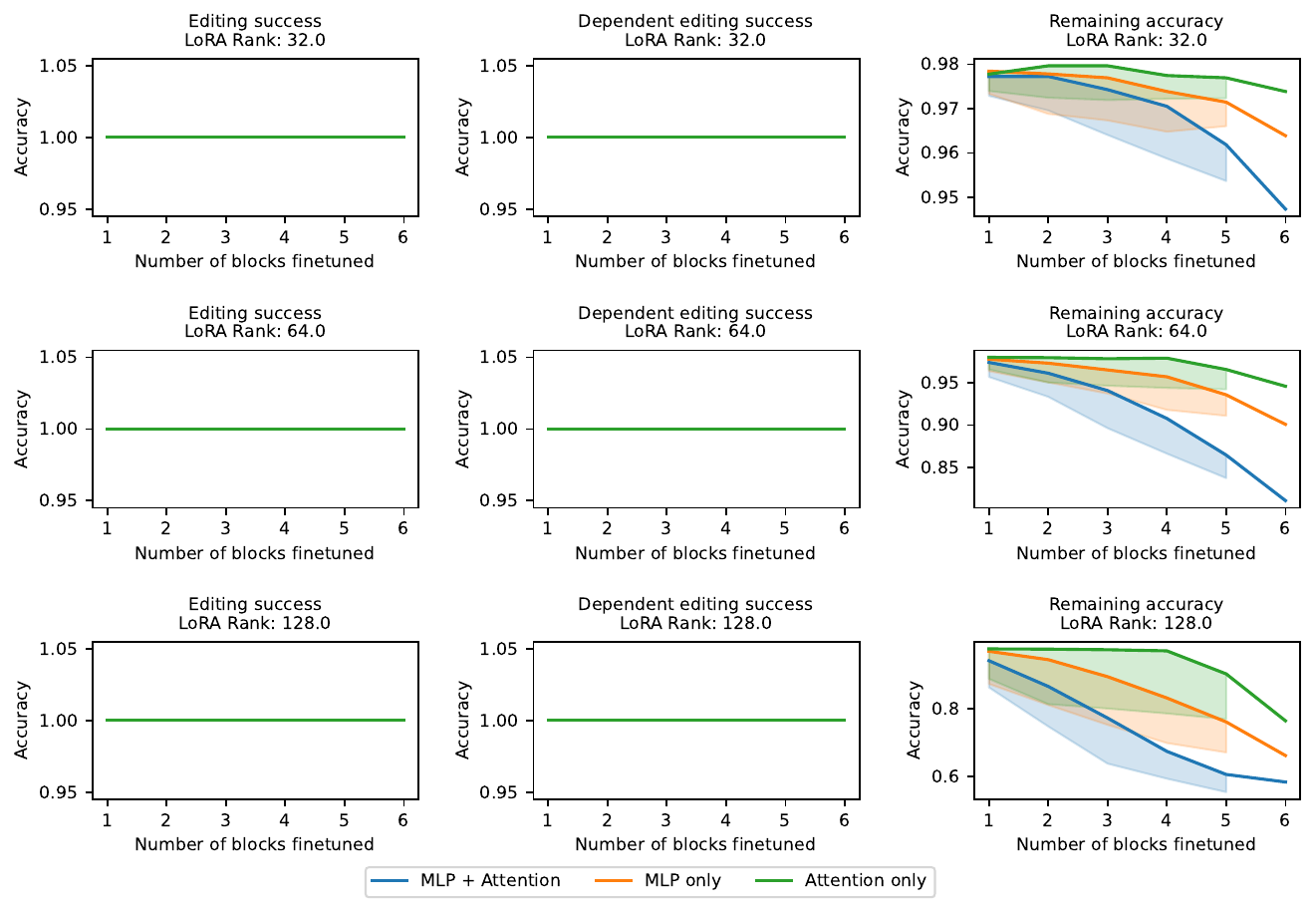}
    \hspace{1em}
    \includegraphics[width=0.45\linewidth]{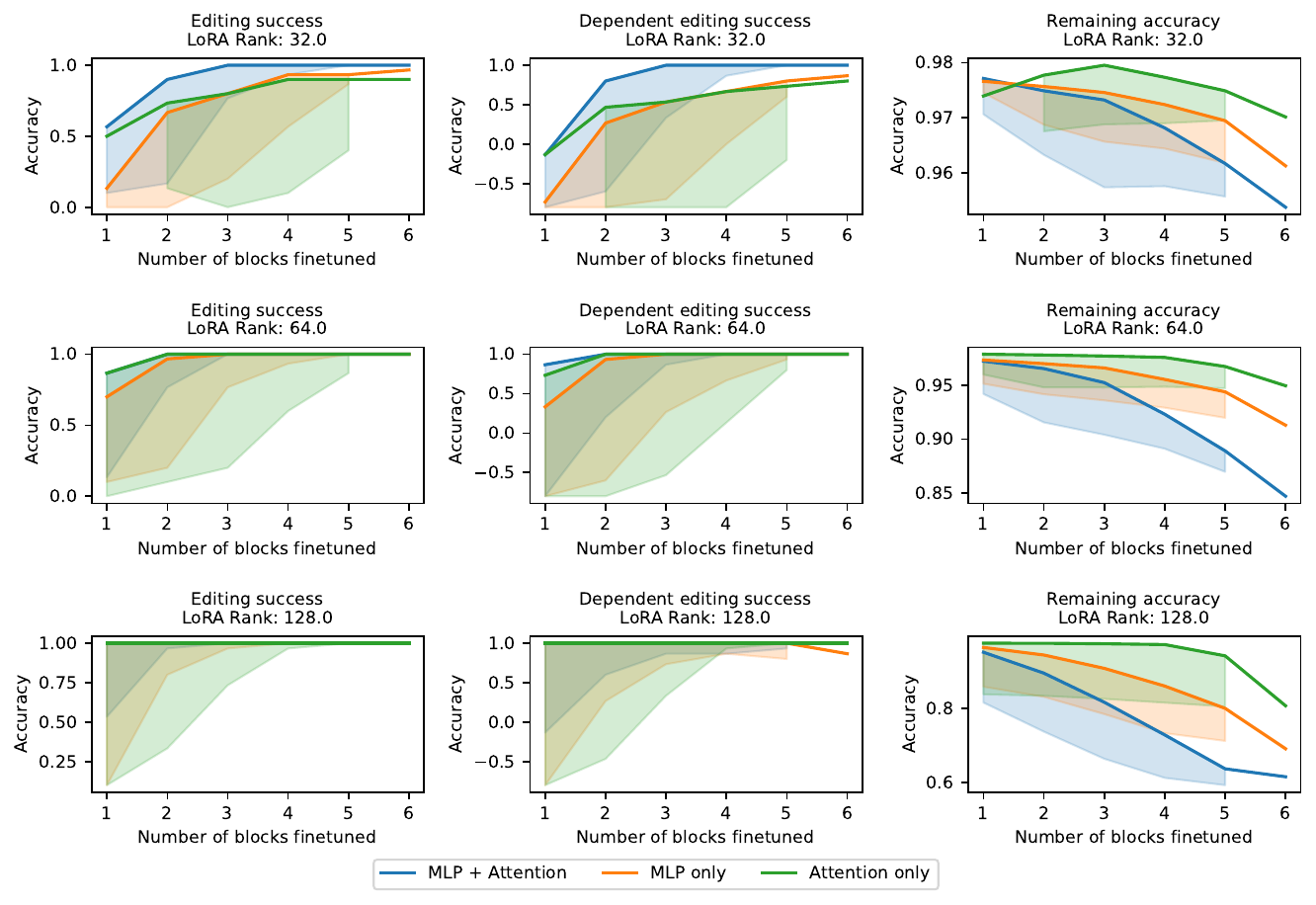} \\
    \vspace{1em}
    \includegraphics[width=0.45\linewidth]{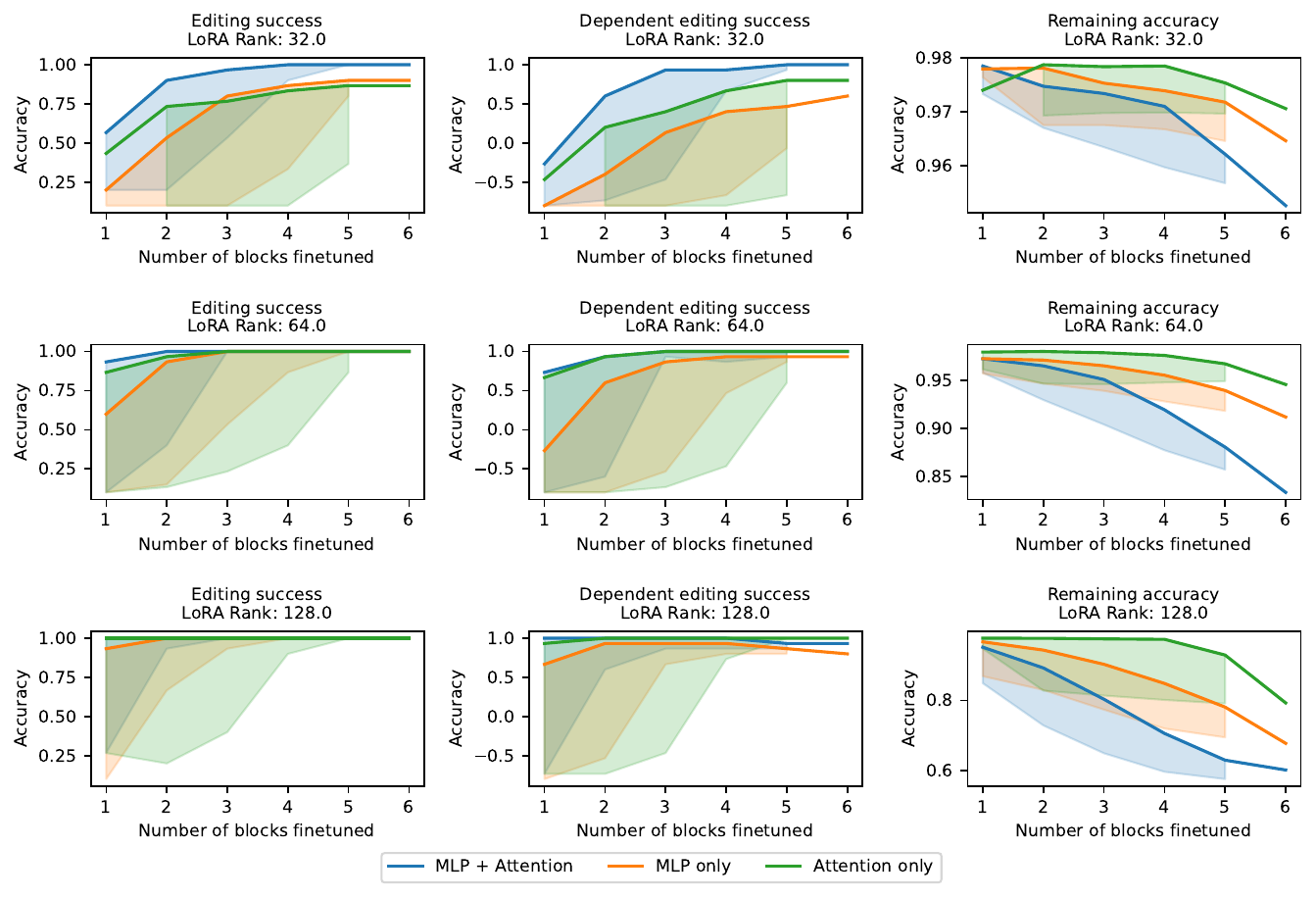}
    \hspace{1em}
    \includegraphics[width=0.45\linewidth]{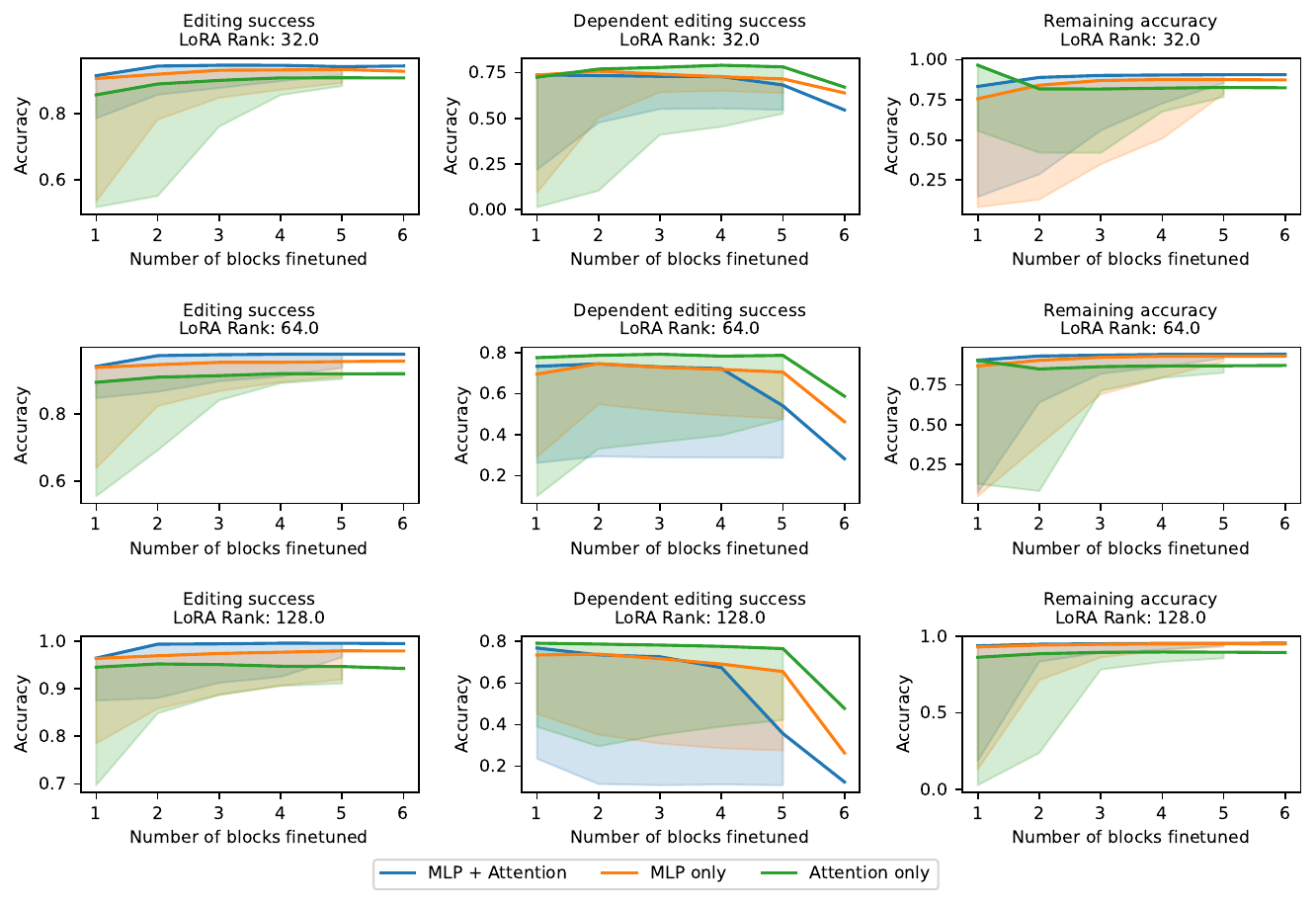}
    \caption{Correlated relationship dataset. Results of editing a single tuple (top, left), making ten identical edits to ten tuples (top, right), making ten different edits (bottom, left), and forgetting a relationship (bottom, right) with LoRA at various ranks, for 75/400/400/2500 training steps. }
    \label{fig:correlated_lora}
\end{figure}

When we consider low-rank finetuning for the correlated relationship dataset, we observe in Figure~\ref{fig:correlated_lora} that, as in the case of full fine-tuning, when making one or several edits, successfully editing the first relationship object generally but not always results in successfully remapping the second relationship object. However, `forgetting' the first relationship can result in either mostly forgetting the second relationship or almost entirely preserving it, depending on the choice of layers. In particular, performing the same correlation analysis as before, we observe from \ref{fig:correlated_dataset_correlations_editing} that fine-tuning the attention layers of the first block or the MLP layers of the last block leads to the loss of editing of the dependent relationship. We also observe that for making several edits, only some of the blocks are positively correlated with making this change. 

\paragraph{Discussion.} Our results in this section show that frequently it is not necessary or desirable to edit every block of an LLM, nor every type of layer within each block. Further, the editing technique can impact the best choice of layer - note that for ROME, it is crucial to edit the first layer, while fine-tuning is more layer-agnostic. These results corroborate recent (concurrent) findings in~\citep{zhu2025newskills}, where the authors also found that fine-tuning either only a subset of the MLP layers or only the attention layers of a multimodal LLM helps prevent catastrophic forgetting. 

\section{Do interpretability techniques help inform our choice?}

\subsection{Activation patching}
We first ask whether we can use the activation patching technique introduced in~\cite{meng2022locating} to select which layers it is best to edit.

To investigate this, we use the activation patching technique described in~\cite{meng2022locating} under the name ``causal tracing''. In this technique, a piece of clean data (a sentence corresponding to a subject, relationship, object tuple) is first run through the model. Then, another subject-relationship pair is run through the model, in particular one with a different object, which we call the "corrupted run". Then, the clean sample is run through the corrupted model again, but this time, the corrupted activations of some part of the model are replaced with the clean activations. We can then look at the values output logits to see where replacing the corrupted activation with the clean one results in moving the clean logit weight toward its original (clean) value. As a technical detail, since the object identifier consists of two tokens, we simply pick cases where the first token of the clean and corrupted object is the same, and look at the second token.

We use the IOI (Indirect Object Identification) metric, which measures the net indirect effect as the logit weight difference between the correct value logit and the corrupted value logit, as follows.
\begin{equation}
        M_f = \frac{(L_{correct, restored}-L_{incorrect, restored})-(L_{correct, corrupted}-L_{incorrect, corrupted})}{(L_{correct, clean}- L_{incorrect,clean})-(L_{correct, corrupted}-L_{incorrect, corrupted})}
    \end{equation}

We collect these measurements at three sites: after each MLP block (post-activation function), after each Attention block, and at the residual stream at the start of each block. We observe that there are no strong differences when we compare the activation patching results for different relationships - in other words, we see no evidence that $\{s,r,o\}$ tuples are in any way partitioned across the layers of the network depending on the specific relationship - even though in this experiment, the objects selected for each relationship for a subject are completely independent of one another. We see the same effect when we look at Attention heads; we further note that the first attention layer and the middle MLP layers seem to be the most crucial: activation patches at the first layer (at the subject token position) and at the last layer (at the final position) result in moving the prediction weight back toward the original, correct prediction. For attention layers, only activation patching at the first block results in moving predictions back to the correct position. However, when we compare these results to our findings in Section~\ref{sec:sub_blocks}, we note that this matches the most effectively edited blocks for ROME, but not for fine-tuning, where the final blocks resulted in the same editing accuracy but better accuracy preservation. For the MLP layer, we observe that the first block is most important at the subject token, with all subsequent blocks being influential at the object token; comparing this observation to the model editing results, we observe that this matches the results for ROME, where the model editing was most successful when applied to the first block, but somewhat overlooks the results for fine-tuning, where the last blocks are the most helpful when the final blocks are the most helpful for maintaining model accuracy when editing the model.

In Appendix~\ref{sec:all_activation_patching}, we show the activation patching results for other relationships, which very strongly resemble the results for the first relationship.

\begin{figure}
    \centering
    \includegraphics[width=0.3\linewidth]{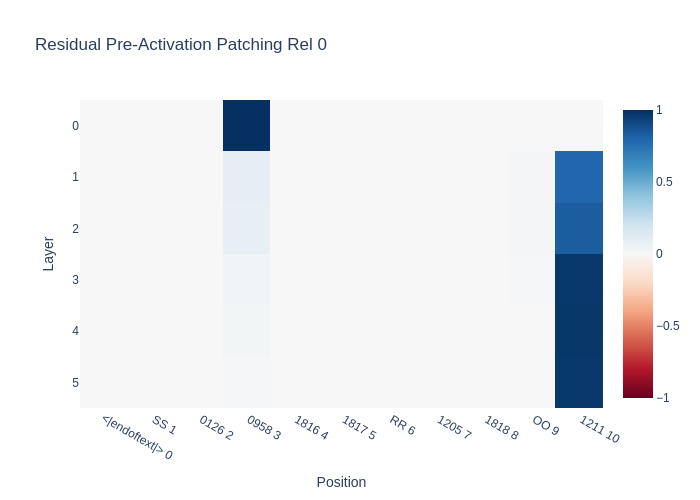}
    \includegraphics[width=0.3\linewidth]{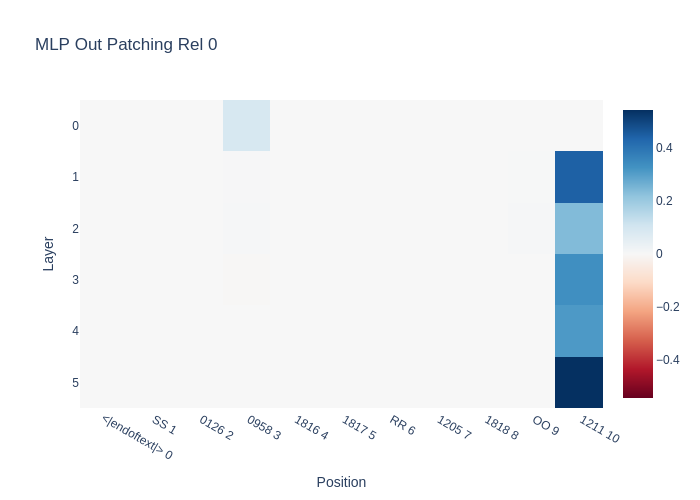}
    \includegraphics[width=0.3\linewidth]{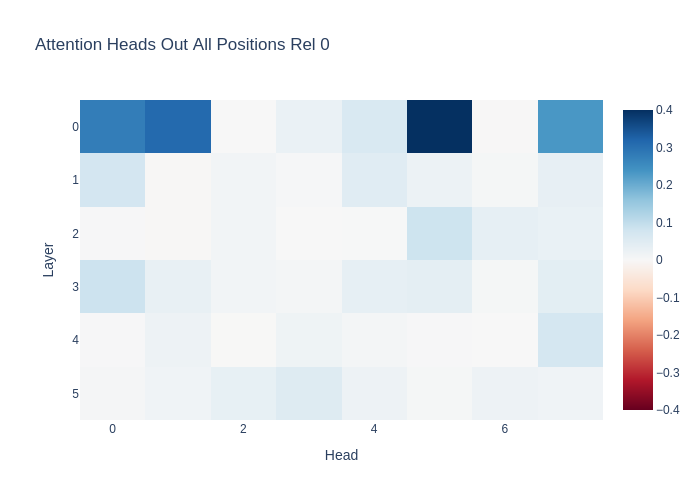}
    \caption{Activation patching IOI metric results for dataset of 120000 $\{s, o, r\}$ tuples}
    \label{fig:mlp_attribution_learning}
\end{figure}

\subsection{Rank estimation}

In this section, we consider whether the effectiveness of the low-rank update matches the intuition we get by considering the estimated rank of the layer movement when doing full finetuning. We estimate the effective rank of the full fine-tuning by performing a singular value decomposition of the matrix $\Delta W:= W_f - W_0$, and measure the rank at which 95\% of the difference is captured.

\begin{figure}
    \centering
    \includegraphics[width=0.24\linewidth]{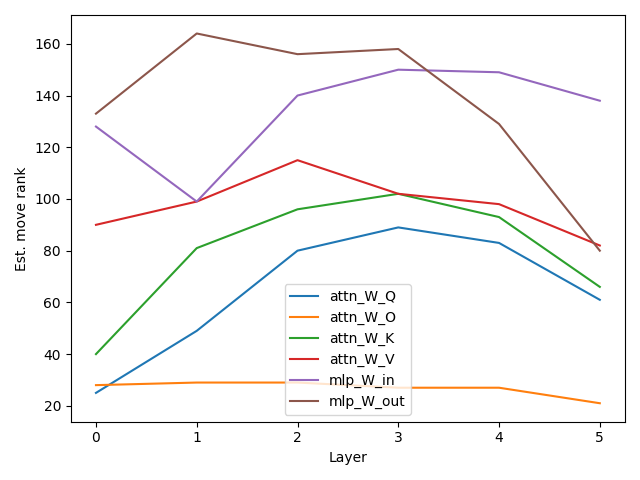}
    \includegraphics[width=0.24\linewidth]{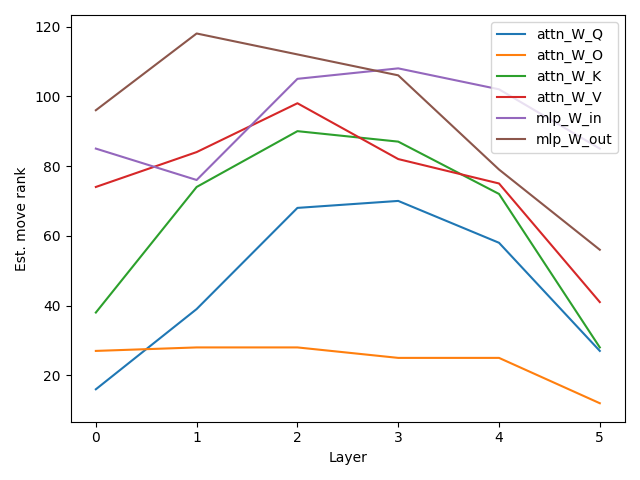}
    \includegraphics[width=0.24\linewidth]{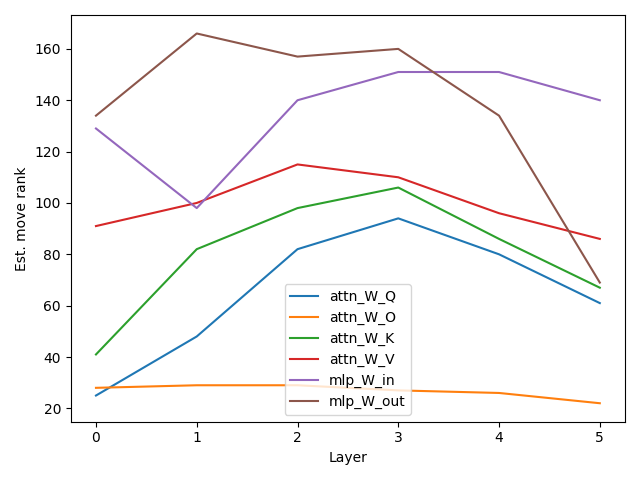}
    \includegraphics[width=0.24\linewidth]{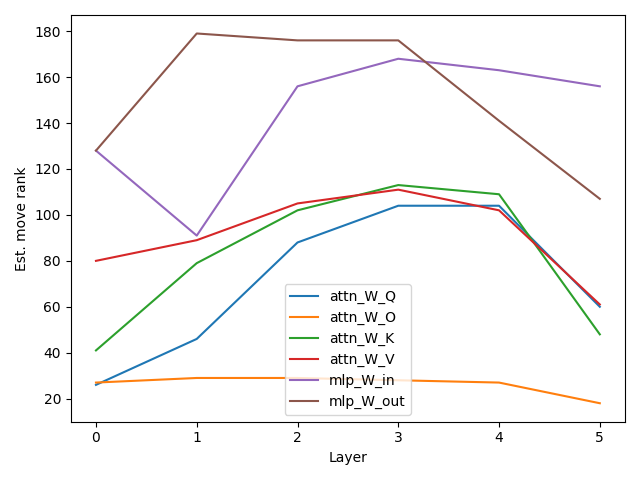}\\
    \includegraphics[width=0.24\linewidth]{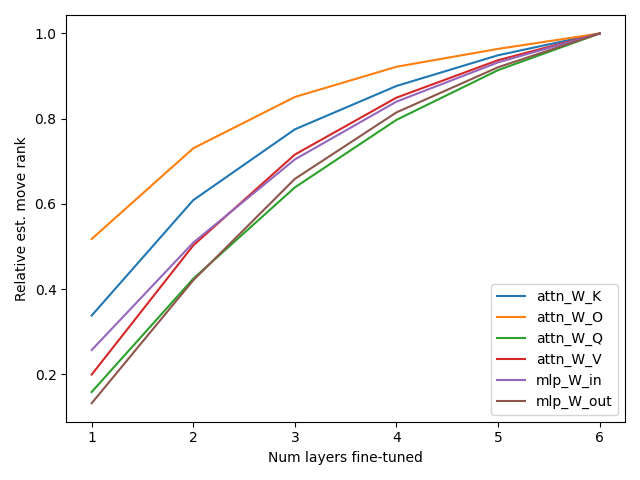}
    \includegraphics[width=0.24\linewidth]{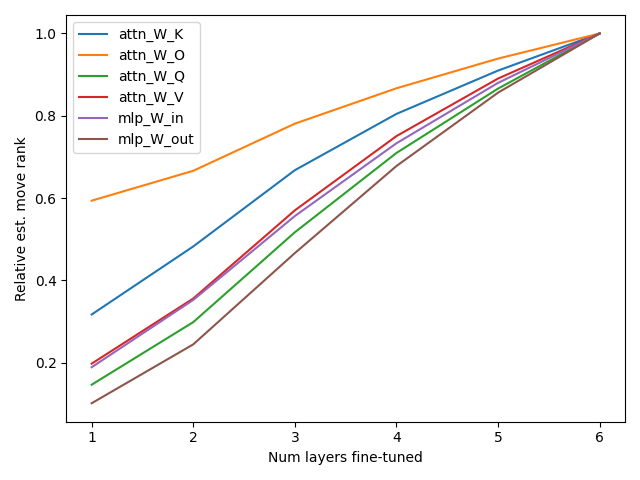}
    \includegraphics[width=0.24\linewidth]{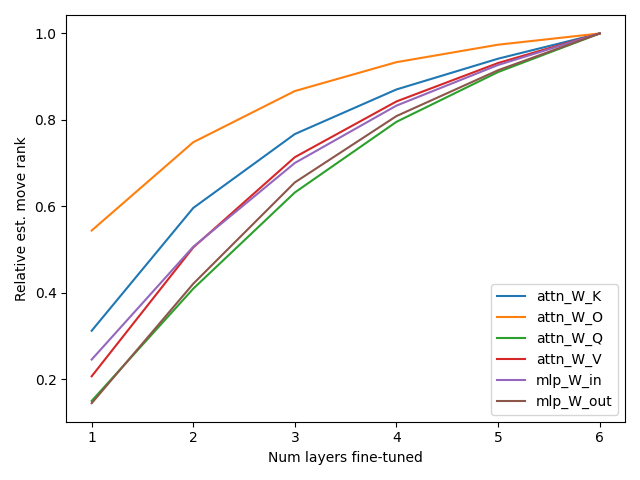}
    \includegraphics[width=0.24\linewidth]{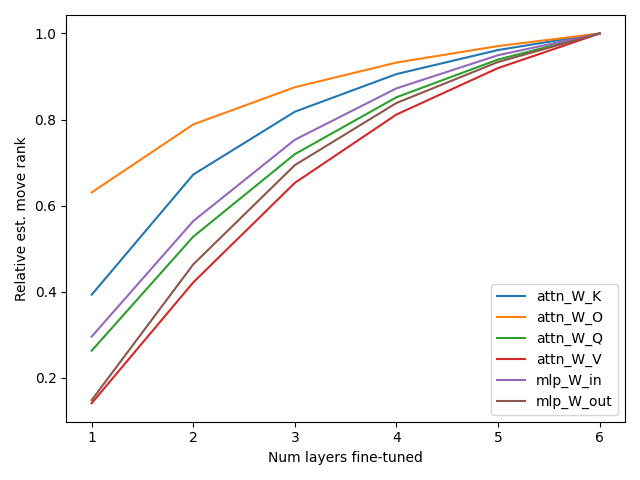}
    \caption{Simple dataset. From left to right - single edit, ten identical edits, ten different edits, forgetting a relationship. (Top) The approximate move rank for components of various blocks, when all layers are fine-tuned with full fine-tuning. (Bottom) The average move rank, relative to the rank when all blocks are fine-tuned, when only some blocks are fine-tuned.}
    \label{fig:move_rank_simple}
\end{figure}

The results of this measurement are shown in Figure~\ref{fig:move_rank_simple}. We observe that when only a single edit is made, the effective rank of the layer weight changes is quite low - under 40 - especially for the MLP layers (recall that the full rank of this model is 256), demonstrating that in this case, the fine-tuning naturally finds a low-rank adaptation of the model. If not all blocks are fine-tuned, the effective rank increases little, up to 27\% higher when only a single block is fine-tuned.

However, when a larger change is made - either ten of the same edit to ten different tuples, or ten different edits, or forgetting a relationship, we observe a vastly different pattern: in all cases, the change becomes much higher rank, especially for the MLP $W_{in}$ layers of the early blocks and all $W_{out}$ layers. Further, restricting the fine-tuned blocks uniformly \textit{lowers} the effective weight change rank, in some cases by a factor of 10. This suggests that given a small amount of layers to fine-tune, the weight adjustments result in small, atomic changes that, in case of making edits, still result in successful model editing; given more `freedom', the changes become more diffused, perhaps slightly over-fitting to the samples drawn from the rest of the data - note that test accuracy from full finetuning does not improve when more blocks are added, past the first two blocks. Recall also that the details of assembling the single-edit fine-tuning dataset and the multiple-edits fine-tuning dataset, and thus the ratios of edited and clean examples, are the same. 

However, when examining the LoRA performance in Figure~\ref{fig:single_lora}, we observe that, when the LoRA rank is 64 or higher, fine-tuning multiple blocks is not necessary to fully achieve the override, while when using a LoRA R of 32, fine-tuning multiple blocks is generally necessary. These observations suggest that the estimated dimension of the weight change during full fine-tuning is not a helpful guide for LoRA rank selection, and also that, in our test scenario, there is a variety of data transformations of various dimensions that result in similar outcomes for model editing.

\section{Conclusion and Limitations}

In this paper, we introduced the Behemoth framework for generating synthetic data for large language models, which uses a custom data generation algorithm and tokenizer to create fully synthetic training data. The relatively compact size of the data and model allows for broad experimentation even with limited computational resources, while the exact known data distribution allows us to more accurately isolate and measure the effects of model editing. We used this framework to evaluate model editing in simple data setups and showed that small hyperparameter or data changes can lead to large effects in the way the model weights are updated and the consequences of these changes. However, we acknowledge that the very simple choice of grammar and data structure is not representative of important attributes of natural-language data. We leave the expansion of the experiments to more complex grammars and data relationships to future work, and hope that this example inspires us and other researchers to continue experimenting with synthetic data.

\section*{Acknowledgements}
EI thanks Weiwei Yang, Janardhan Kulkani, and Kate Lytvynets for their advice and support in developing an earlier version of the Behemoth library. This research was supported by the Scientific Service Units (SSU) of IST Austria through resources provided by Scientific Computing (SciComp). EI was supported in part by the FWF DK VGSCO, grant agreement number W1260-N35.

\clearpage

\bibliography{references.bib}

@inproceedings{Sun2023Wanda,
  title={A Simple and Effective Pruning Approach for Large Language Models},
  author={Mingjie Sun and Zhuang Liu and Anna Bair and J. Zico Kolter},
  booktitle={International Conference on Learning Representations (ICLR)},
  year={2023},
}

@inproceedings{Qi2024GPTJailbreak,
  title={Fine-tuning Aligned Language Models Compromises Safety, Even When Users Do Not Intend To!},
  author={Xiangyu Qi and Yi Zeng and Tinghao Xie and Pin-Yu Chen and Ruoxi Jia and Prateek Mittal and Peter Henderson},
  booktitle={International Conference on Learning Representations (ICLR)},
  year={2024},
}

@misc{Wei2024Brittle,
  title={Assessing the Brittleness of Safety Alignment via Pruning and Low-Rank Modifications},
  author={Boyi Wei and Kaixuan Huang and Yangsibo Huang and Tinghao Xie and Xiangyu Qi and Mengzhou Xia and Prateek Mittal and Mengdi Wang and Peter Henderson},
  year={2024},
  howpublished={arXiv}
}

@misc{Biderman2024lora,
    title={LoRA Learns Less and Forgets Less},
    author={Dan Biderman and Jose Gonzalez Ortiz and Jacob Portes and Mansheej Paul and Philip Greengard and Connor Jennings and Daniel King and Sam Havens and Vitaliy Chiley and Jonathan Frankle and Cody Blakeney and John P. Cunningham},
    year={2024},
    eprint={2405.09673},
    howpublished={arXiv},
}

@misc{Templeton2024ClaudeMonosemanticity,
    title={Scaling Monosemanticity: Extracting Interpretable Features from Claude 3 Sonnet},
    author={Adly Templeton and Tom Conerly and Jonathan Marcus and Jack Lindsey and Trenton Bricken and Brian Chen and Adam Pearce and Craig Citro and Emmanuel Ameisen and Andy Jones and Hoagy Cunningham and Nicholas L Turner and Callum McDougall and Monte MacDiarmid and Alex Tamkin and Esin Durmus and Tristan Hume and Francesco Mosconi and C. Daniel Freeman and Theodore R. Sumers and Edward Rees and Joshua Batson and Adam Jermyn and Shan Carter and Chris Olah and Tom Henighan},
    year={2024},
    howpublished={https://transformer-circuits.pub/2024/scaling-monosemanticity/index.html},
}

@inproceedings{meng2023memit,
  title={Mass Editing Memory in a Transformer},
  author={Kevin Meng and Sen Sharma, Arnab and Alex Andonian and Yonatan Belinkov and David Bau},
  booktitle={ICLR},
  year={2023}
}

@misc{Bricken2024ToyMonosemanticity,
    title={Towards Monosemanticity: Decomposing Language Models With Dictionary Learning},
    author={Trenton Bricken and Adly Templeton and Joshua Batson and Brian Chen and Adam Jermyn and Tom Conerly and Nicholas L Turner and Cem Anil and Carson Denison and Amanda Askell and Robert Lasenby and Yifan Wu and Shauna Kravec and Nicholas Schiefer and Tim Maxwell and Nicholas Joseph and Alex Tamkin and Karina Nguyen and Brayden McLean and Josiah E Burke and Tristan Hume and Shan Carter and Tom Henighan and Chris Olah},
    year={2024},
    howpublished={https://transformer-circuits.pub/2024/scaling-monosemanticity/index.html},
}

@misc{allenzhu2024physics,
      title={Physics of Language Models: Part 3.3, Knowledge Capacity Scaling Laws}, 
      author={Zeyuan Allen-Zhu and Yuanzhi Li},
      year={2024},
      eprint={2404.05405},
}

@misc{allenzhu2023physics,
      title={Physics of Language Models: Part 3.1, Knowledge Storage and Extraction}, 
      author={Zeyuan Allen-Zhu and Yuanzhi Li},
      year={2023},
      eprint={2309.14316},
}

@misc{allenzhu2024physics1,
      title={Physics of Language Models: Part 1, Learning Hierarchical Language Structures}, 
      author={Zeyuan Allen-Zhu and Yuanzhi Li},
      year={2024},
      eprint={2305.13673},
}

@inproceedings{lee2018snip,
  title={SNIP: Single-shot network pruning based on connection sensitivity},
  author={Lee, Namhoon and Ajanthan, Thalaiyasingam and Torr, Philip HS},
  booktitle={ICLR},
  year={2019},
}

@misc{eldan2023whosHarryPotter,
      title={Who's Harry Potter? Approximate Unlearning in LLMs}, 
      author={Ronen Eldan and Mark Russinovich},
      year={2023},
      eprint={2310.02238},
      howpublished={arXiv},
}

@misc{Lynch2024EightMethodsRobustUnlearning,
  title={Eight Methods to Evaluate Robust Unlearning in LLMs},
  author={Aengus Lynch and Phillip Guo and Aidan Ewart and Stephen Casper and Dylan Hadfield-Menell},
      howpublished={arXiv},
  year={2024},
  eprint={2402.16835},
}

@misc{Maini2024TOFU,
  title={TOFU: A Task of Fictitious Unlearning for LLMs},
  author={Pratyush Maini and Zhili Feng and Avi Schwarzschild and Zachary Chase Lipton and J. Zico Kolter},
  howpublished={ArXiv},
  year={2024},
  volume={abs/2401.06121},
}

@inproceedings{jain2023mechanistically,
      title={Mechanistically analyzing the effects of fine-tuning on procedurally defined tasks}, 
      author={Samyak Jain and Robert Kirk and Ekdeep Singh Lubana and Robert P. Dick and Hidenori Tanaka and Edward Grefenstette and Tim Rocktäschel and David Scott Krueger},
      year={2024},
      booktitle={ICLR},
}

@inproceedings{kotha2024understandingCatastrophicForgetting,
      title={Understanding Catastrophic Forgetting in Language Models via Implicit Inference}, 
      author={Suhas Kotha and Jacob Mitchell Springer and Aditi Raghunathan},
      year={2024},
      booktitle={ICLR},
}

@misc{makelov2024principled,
      title={Towards Principled Evaluations of Sparse Autoencoders for Interpretability and Control}, 
      author={Aleksandar Makelov and George Lange and Neel Nanda},
      year={2024},
      eprint={2405.08366},
      howpublished={ArXiv},
}

@misc{wang2022interpretability,
      title={Interpretability in the Wild: a Circuit for Indirect Object Identification in GPT-2 small}, 
      author={Kevin Wang and Alexandre Variengien and Arthur Conmy and Buck Shlegeris and Jacob Steinhardt},
      year={2022},
      eprint={2211.00593},
      howpublished={ArXiv},
}

@inproceedings{meng2022locating,
  title={Locating and Editing Factual Associations in {GPT}},
  author={Kevin Meng and David Bau and Alex Andonian and Yonatan Belinkov},
  booktitle={NeurIPS},
  year={2022},
}

@misc{hartmann2023sok,
      title={SoK: Memorization in General-Purpose Large Language Models}, 
      author={Valentin Hartmann and Anshuman Suri and Vincent Bindschaedler and David Evans and Shruti Tople and Robert West},
      year={2023},
      eprint={2310.18362},
      howpublished={arXiv},
}

@misc{qi2024safetyalignment,
      title={Safety Alignment Should Be Made More Than Just a Few Tokens Deep}, 
      author={Xiangyu Qi and Ashwinee Panda and Kaifeng Lyu and Xiao Ma and Subhrajit Roy and Ahmad Beirami and Prateek Mittal and Peter Henderson},
      year={2024},
      eprint={2406.05946},
}

@misc{zhu2025newskills,
      title={How to Teach Large Multimodal Models New Skills}, 
      author={Zhen Zhu and Yiming Gong and Yao Xiao and Yaoyao Liu and Derek Hoiem},
      year={2025},
      eprint={2510.08564},
}

@inproceedings{Hong2024DissectingFU,
  title={Dissecting Fine-Tuning Unlearning in Large Language Models},
  author={Yihuai Hong and Yuelin Zou and Lijie Hu and Ziqian Zeng and Di Wang and Haiqin Yang},
  booktitle={Conference on Empirical Methods in Natural Language Processing},
  year={2025},
}

@inproceedings{Joshi2024TowardsRE,
  title={Towards Robust Evaluation of Unlearning in LLMs via Data Transformations},
  author={Abhinav Joshi and Shaswati Saha and Divyaksh Shukla and Sriram Vema and Harsh Jhamtani and Manas Gaur and Ashutosh Modi},
  booktitle={Conference on Empirical Methods in Natural Language Processing},
  year={2024},
}

@article{Krishnan2025NotAD,
  title={Not All Data Are Unlearned Equally},
  author={Aravind Krishnan and Siva Reddy and Marius Mosbach},
  journal={ArXiv},
  year={2025},
  volume={abs/2504.05058},
}

@article{Wu2024EvaluatingDU,
  title={Evaluating Deep Unlearning in Large Language Models},
  author={Ruihan Wu and Chhavi Yadav and Russ Salakhutdinov and Kamalika Chaudhuri},
  journal={ArXiv},
  year={2024},
  volume={abs/2410.15153},
}

@misc{litgpt-2023,
  author       = {{Lightning AI}},
  title        = {LitGPT},
  howpublished = {\url{https://github.com/Lightning-AI/litgpt}},
  year         = {2023},
}

@article{Morris2025HowMD,
  title={How much do language models memorize?},
  author={John X. Morris and Chawin Sitawarin and Chuan Guo and Narine Kokhlikyan and G. Edward Suh and Alexander M. Rush and Kamalika Chaudhuri and Saeed Mahloujifar and Fair at Meta and Google Deepmind},
  journal={ArXiv},
  year={2025},
  volume={abs/2505.24832},
}

\appendix

\clearpage

\section{Activation patching results for all relationships}
\label{sec:all_activation_patching}

In this section, we present activation patching results for all six relationships for the simple dataset. We observe that there is very little difference between the activation patching results, regardless of the relationship, demonstrating that the important points of the data are stored in the same parts of the network, accounting for the inefficient data storage and, therefore, the lower bits-per-parameter ratios.

\begin{figure}
    \centering
    \includegraphics[width=0.3\linewidth]{graphics/activation_patching/resid_pre_activation_IOI_rel0_combined.png}
    \includegraphics[width=0.3\linewidth]{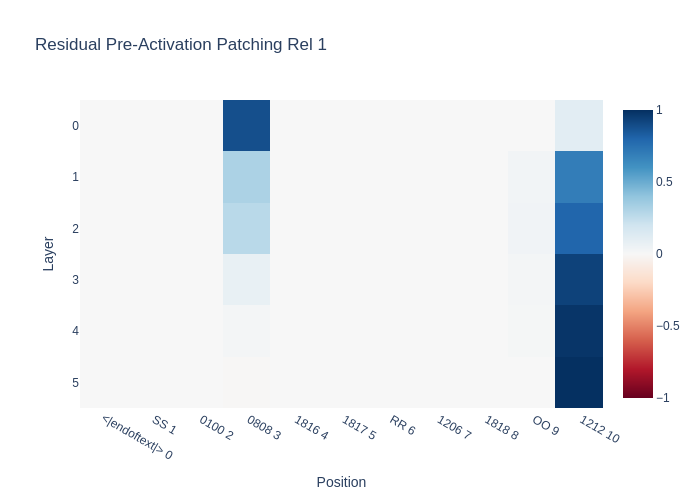}
    \includegraphics[width=0.3\linewidth]{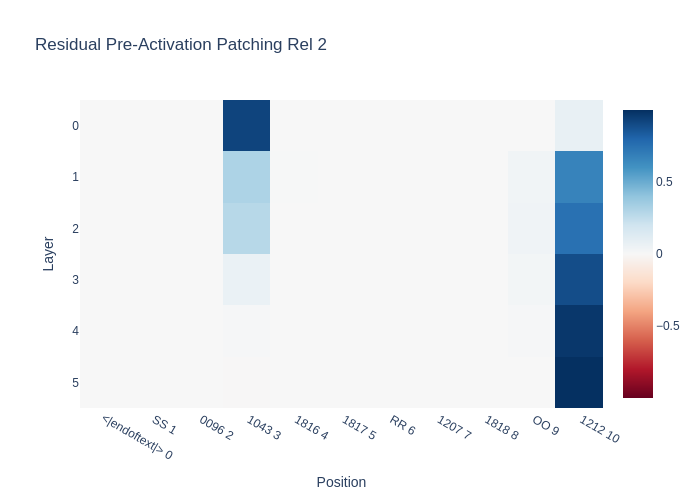} \\
    \includegraphics[width=0.3\linewidth]{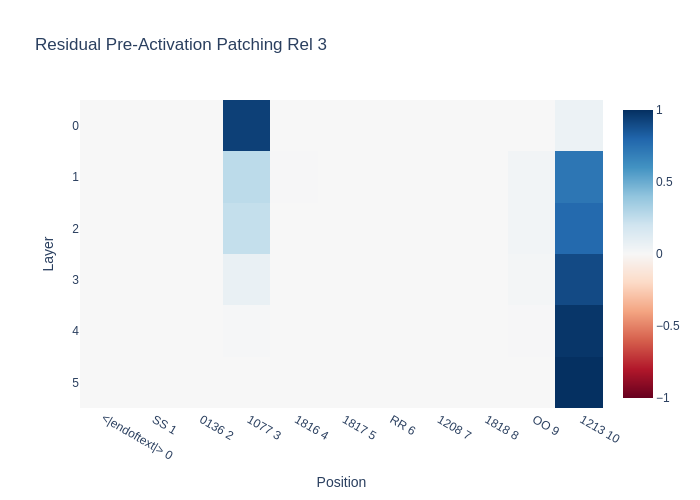}
    \includegraphics[width=0.3\linewidth]{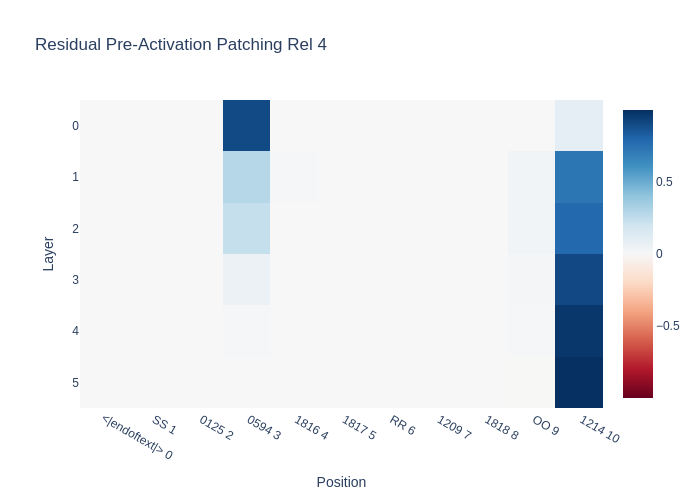}
    \includegraphics[width=0.3\linewidth]{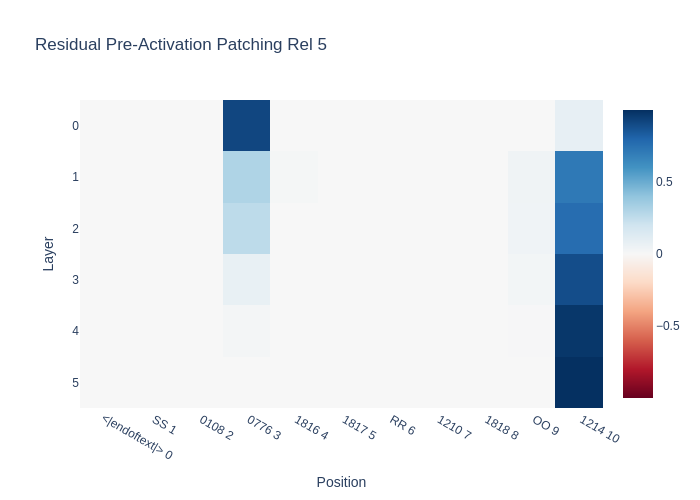} 
    \caption{Simple dataset. Activation patching IOI metric results for all relationship $\{s, o, r\}$ tuples, residual pre activation.}
    \label{fig:graphics/activation_patching_allrel_resid}
\end{figure}

\begin{figure}
    \centering
    \includegraphics[width=0.3\linewidth]{graphics/activation_patching/mlp_out_IOI_rel0_combined.png}
    \includegraphics[width=0.3\linewidth]{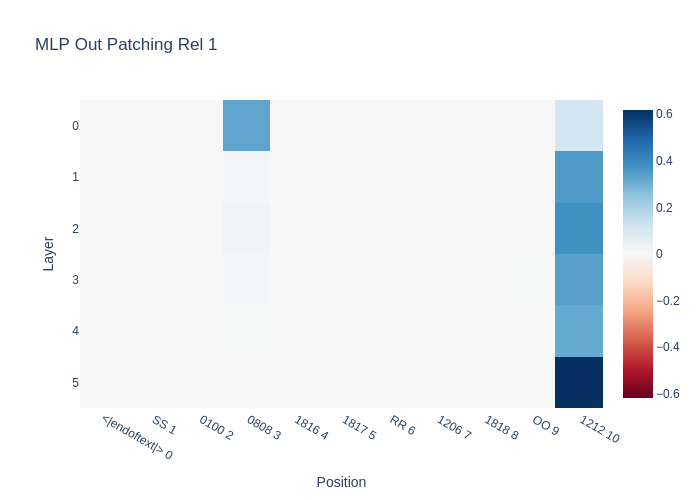}
    \includegraphics[width=0.3\linewidth]{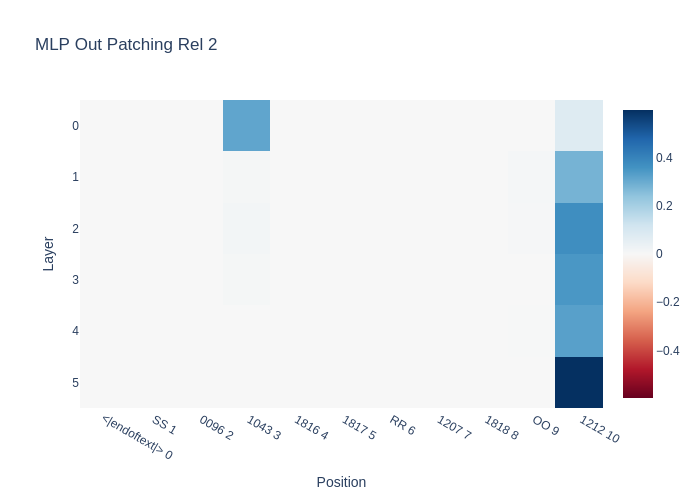}
    \\
    \includegraphics[width=0.3\linewidth]{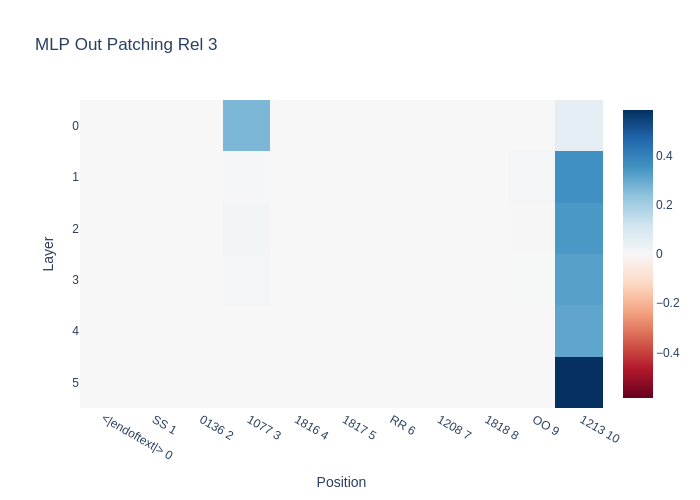}
    \includegraphics[width=0.3\linewidth]{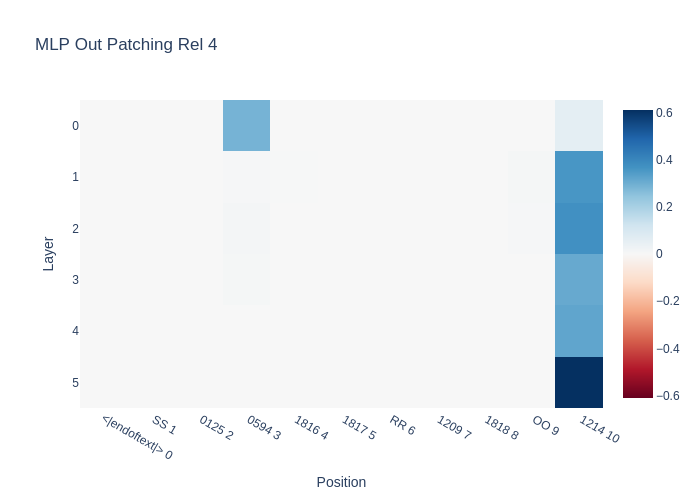}
    \includegraphics[width=0.3\linewidth]{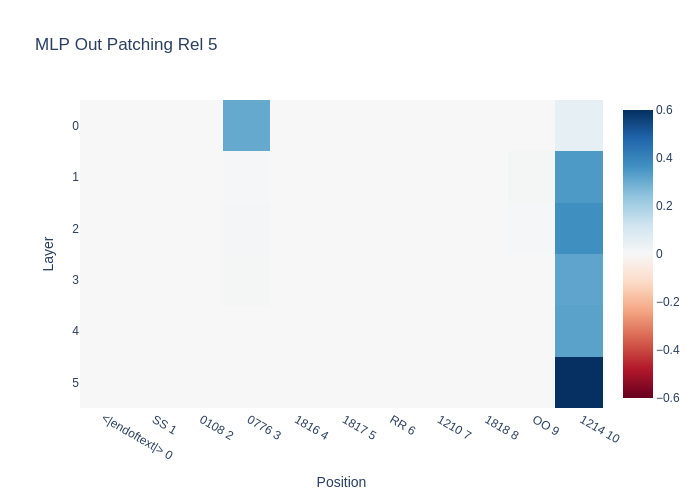}
    \caption{Simple dataset. Activation patching IOI metric results for all relationship $\{s, o, r\}$ tuples, MLP.}
    \label{fig:graphics/activation_patching_allrel_mlp}
\end{figure}

\begin{figure}
    \centering
    \includegraphics[width=0.3\linewidth]{graphics/activation_patching/attn_all_positions_IOI_rel0_combined.png}
    \includegraphics[width=0.3\linewidth]{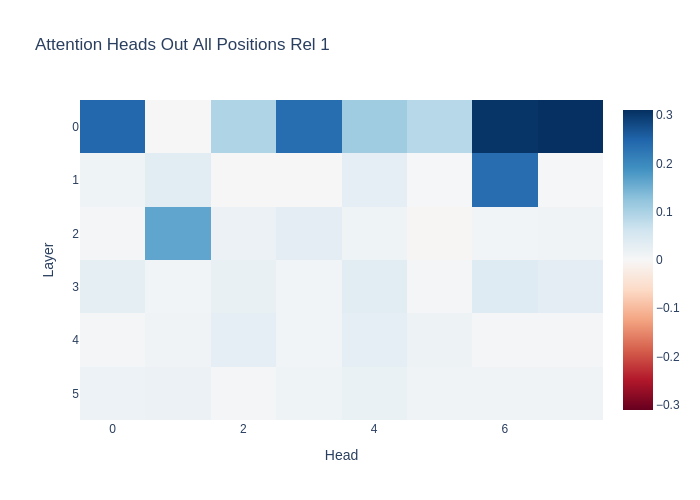}
    \includegraphics[width=0.3\linewidth]{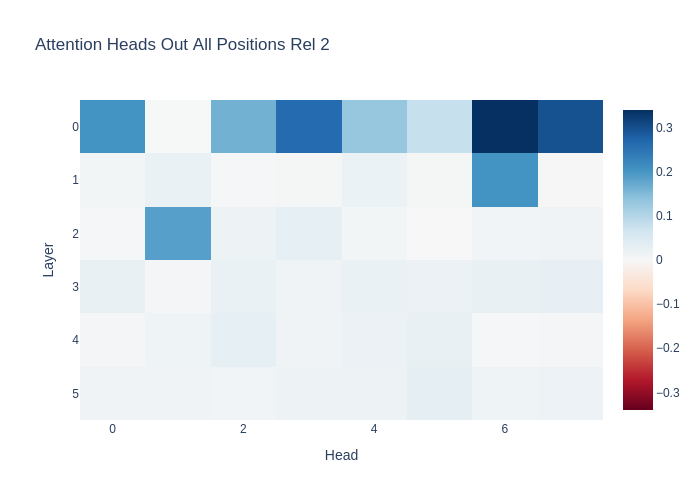}
    \\
    \includegraphics[width=0.3\linewidth]{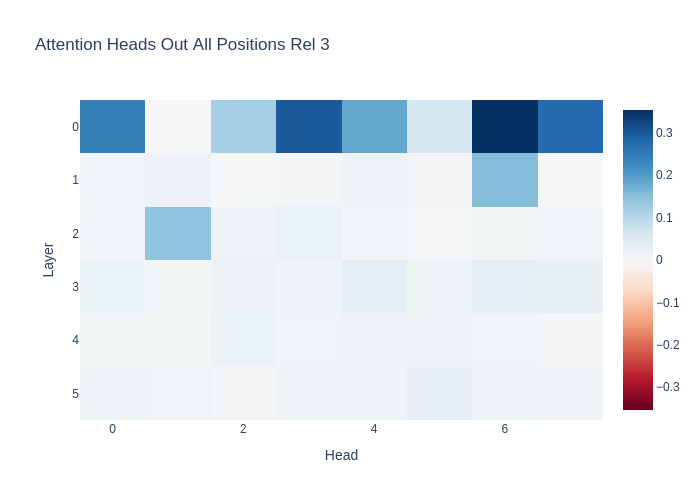}
    \includegraphics[width=0.3\linewidth]{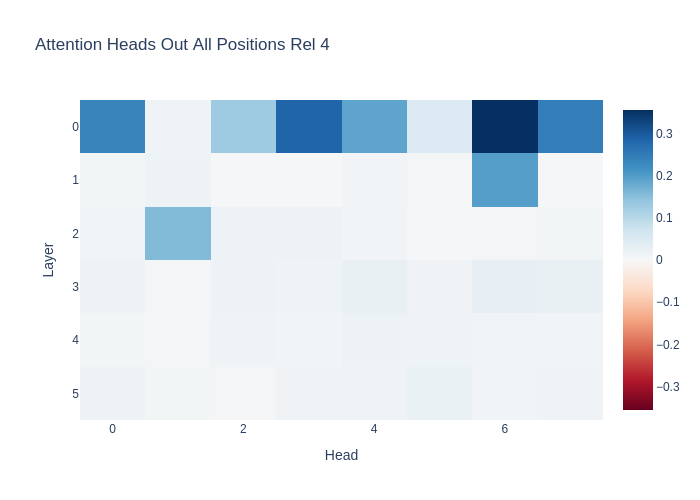}
    \includegraphics[width=0.3\linewidth]{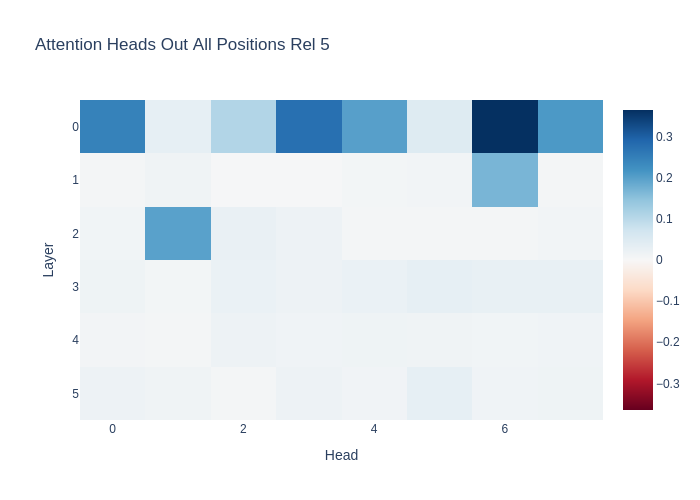}
    \caption{Simple dataset. Activation patching IOI metric results for all relationship $\{s, o, r\}$ tuples, attention all positions.}
    \label{fig:graphics/activation_patching_allrel_attn}
\end{figure}

\section{Performance of individual fine-tuned models (full fine-tuning}

\begin{figure}
    \centering
    \includegraphics[width=0.95\linewidth]{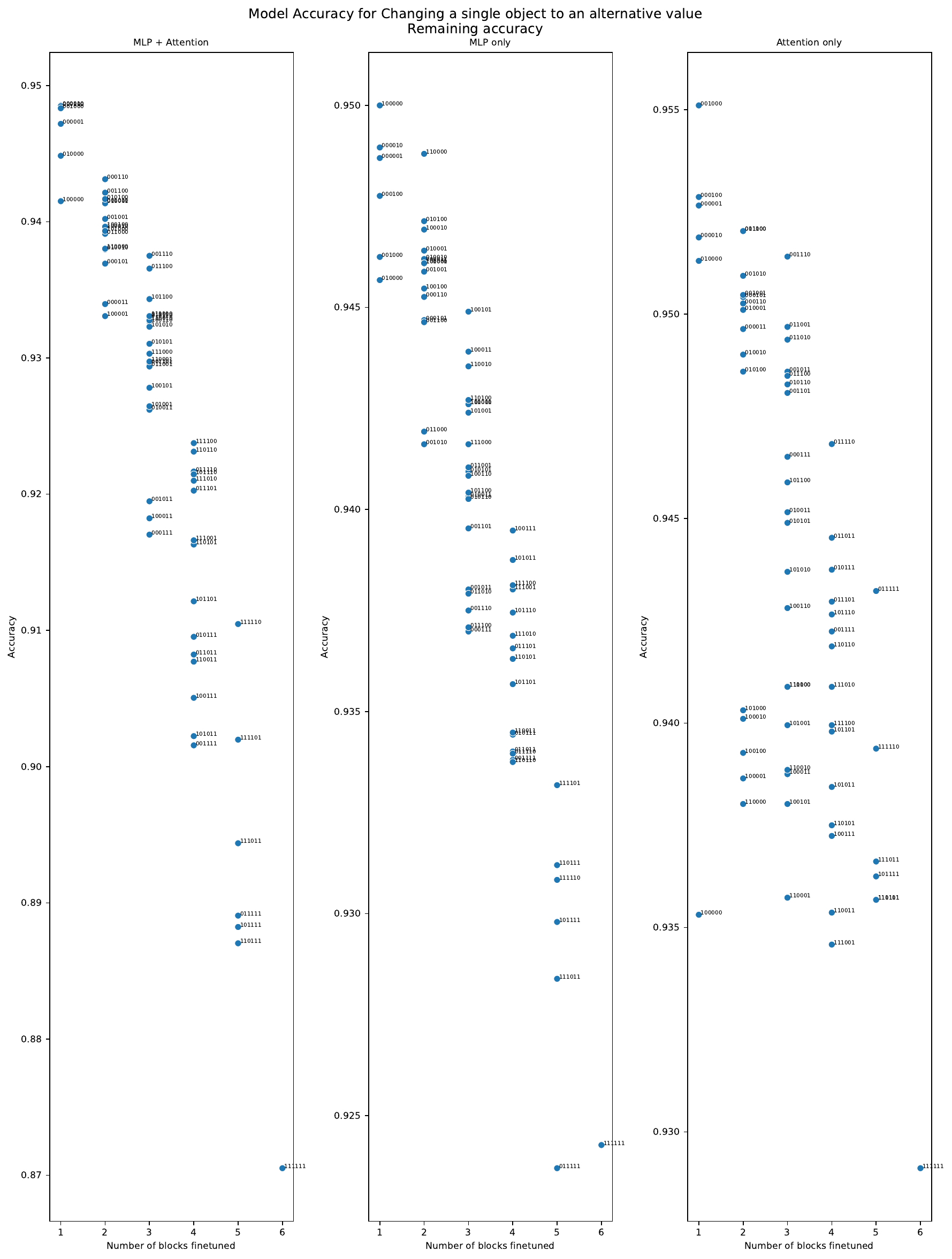}
    \caption{Simple dataset, 1 overrides, remaining data accuracy after full finetuning.}
    \label{fig:simple_1overrides_remainingacc_pointplot}
\end{figure}

\begin{figure}
    \centering
    \includegraphics[width=0.95\linewidth]{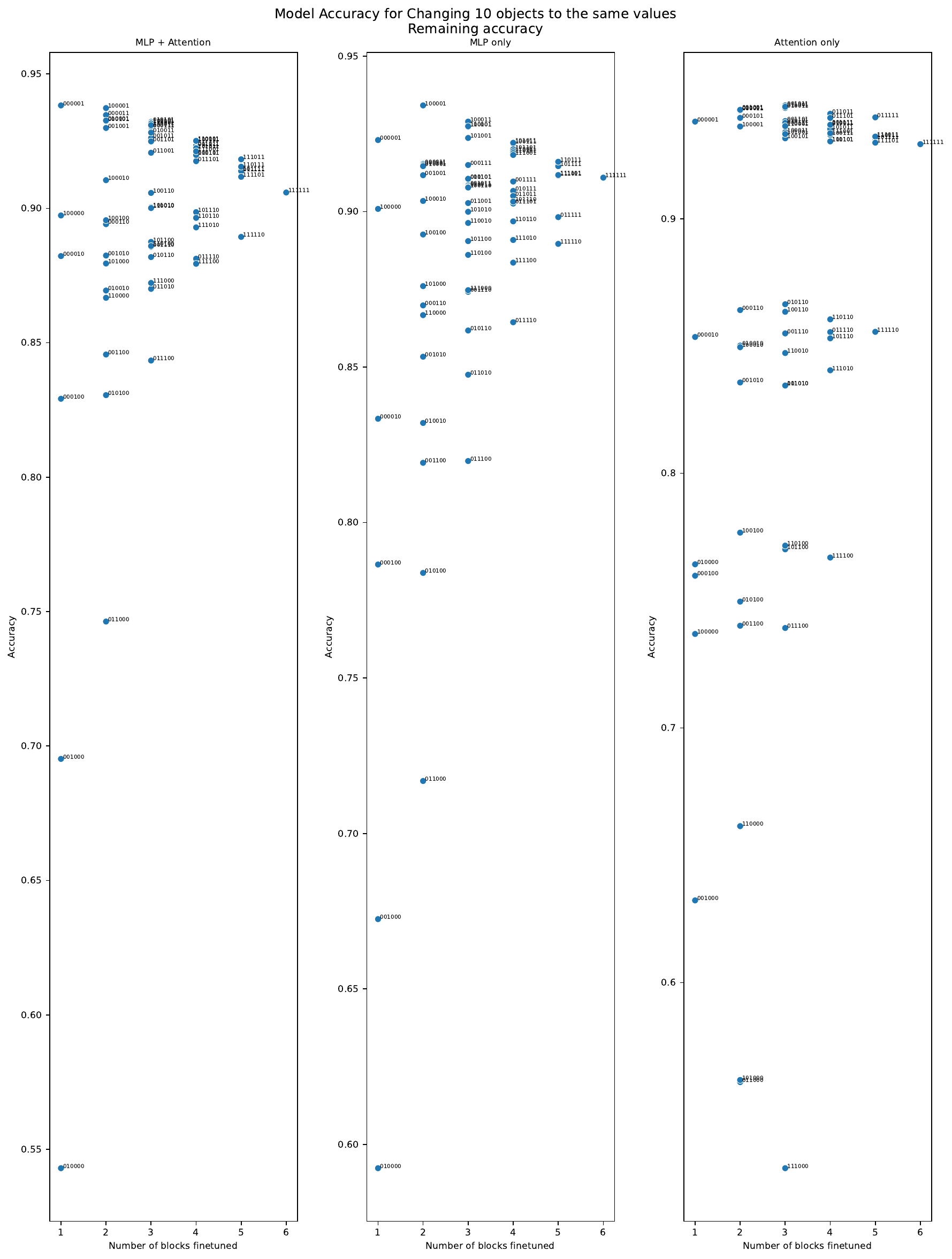}
    \caption{Simple dataset, 10 of the same overrides, remaining data accuracy after full finetuning.}
    \label{fig:simple_10overrides_allsame_remainingacc_pointplot}
\end{figure}

\begin{figure}
    \centering
    \includegraphics[width=0.95\linewidth]{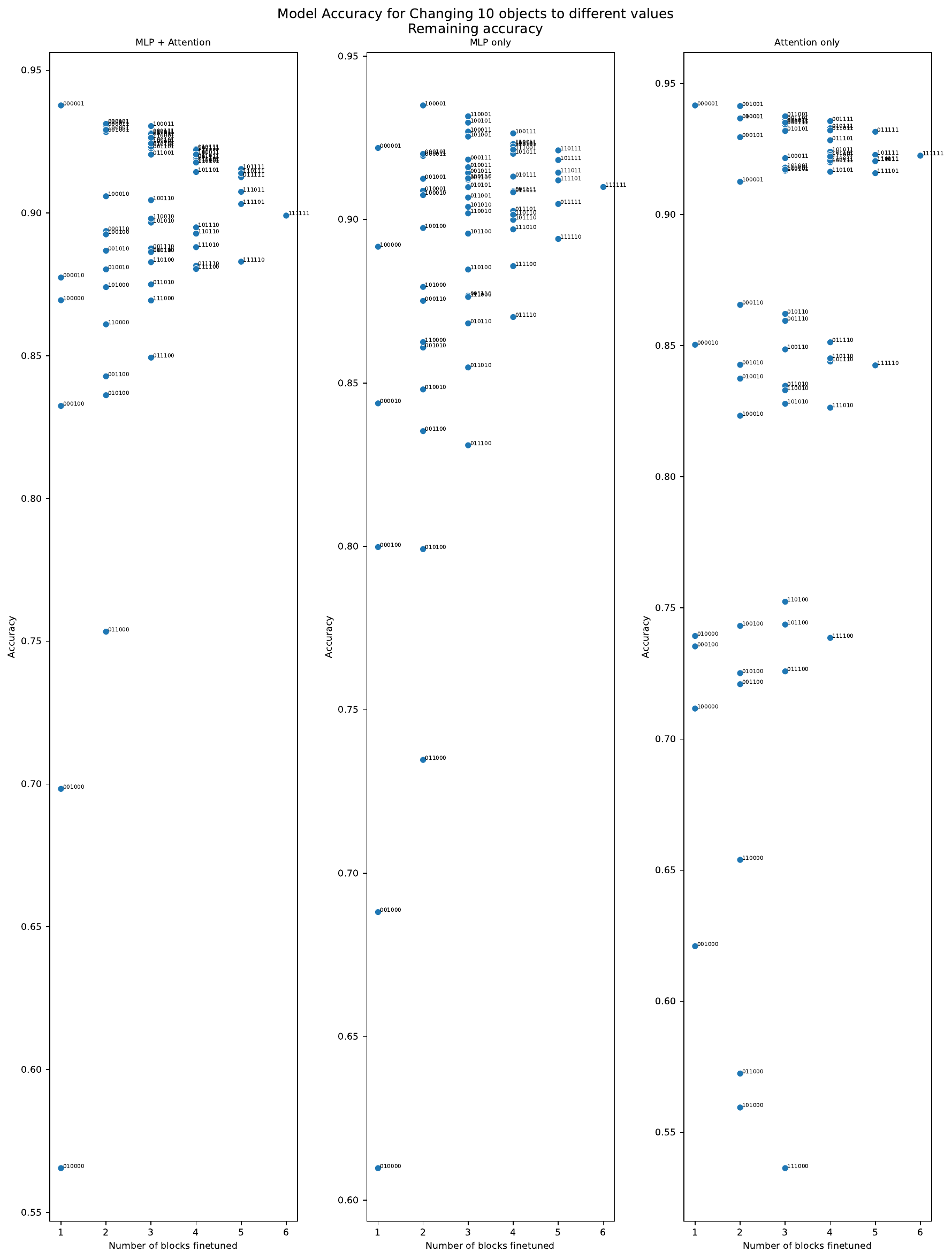}
    \caption{Simple dataset, 10 overrides, remaining data accuracy after full finetuning.}
    \label{fig:simple_10overrides_remainingacc_pointplot}
\end{figure}

\begin{figure}
    \centering
    \includegraphics[width=0.95\linewidth]{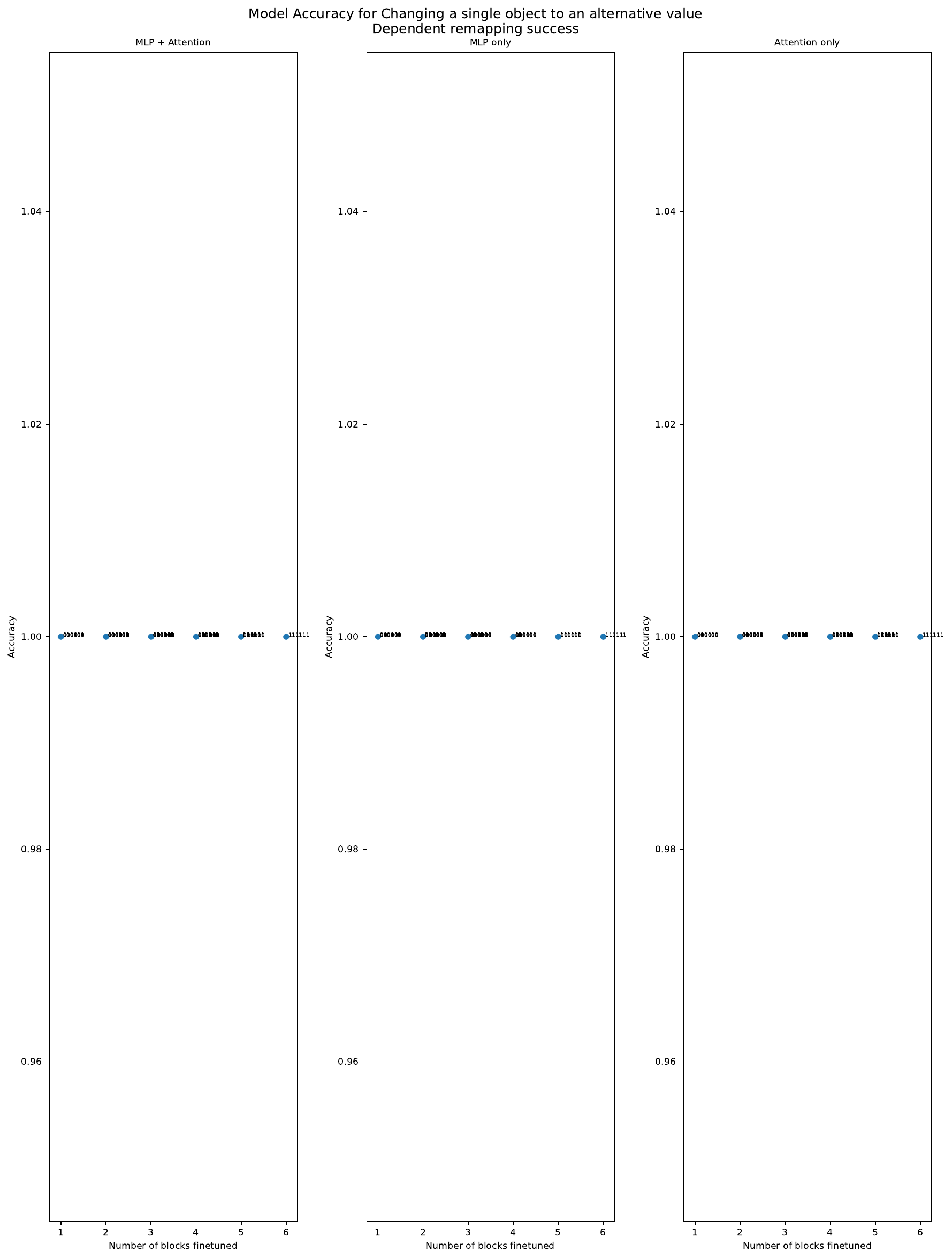}
    \caption{Correlated dataset, 1 override, remaining data accuracy after full finetuning.}
    \label{fig:correlated_1overrides_dependentremapping_pointplot}
\end{figure}

\begin{figure}
    \centering
    \includegraphics[width=0.95\linewidth]{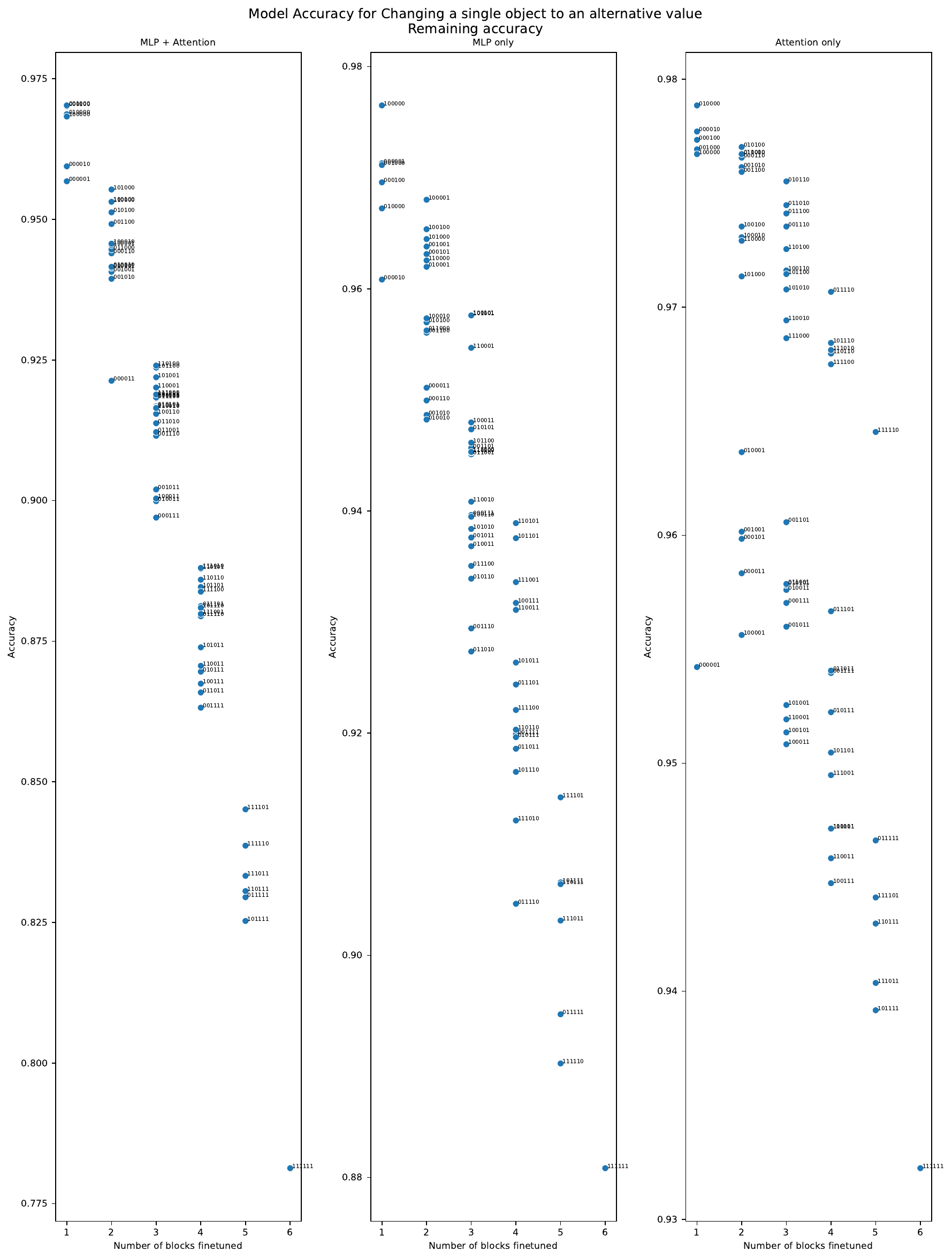}
    \caption{Correlated dataset, 1 override, remaining data accuracy after full finetuning.}
    \label{fig:correlated_1verrides_remainingacc_pointplot}
\end{figure}

\begin{figure}
    \centering
    \includegraphics[width=0.95\linewidth]{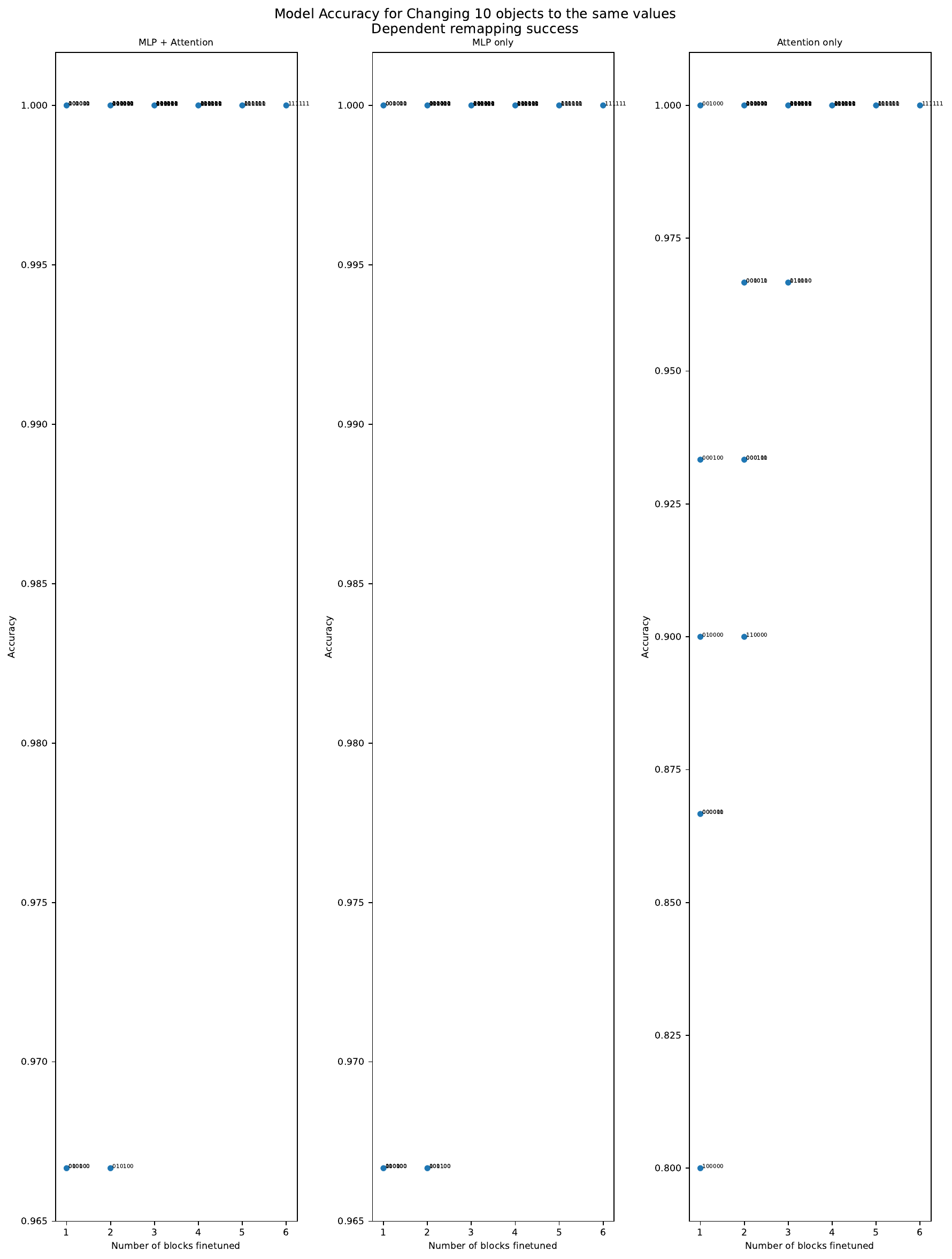}
    \caption{Correlated dataset, 10 of the same override, remaining data accuracy after full finetuning.}
    \label{fig:correlated_10overrides_allsame_dependentremapping_pointplot}
\end{figure}

\begin{figure}
    \centering
    \includegraphics[width=0.95\linewidth]{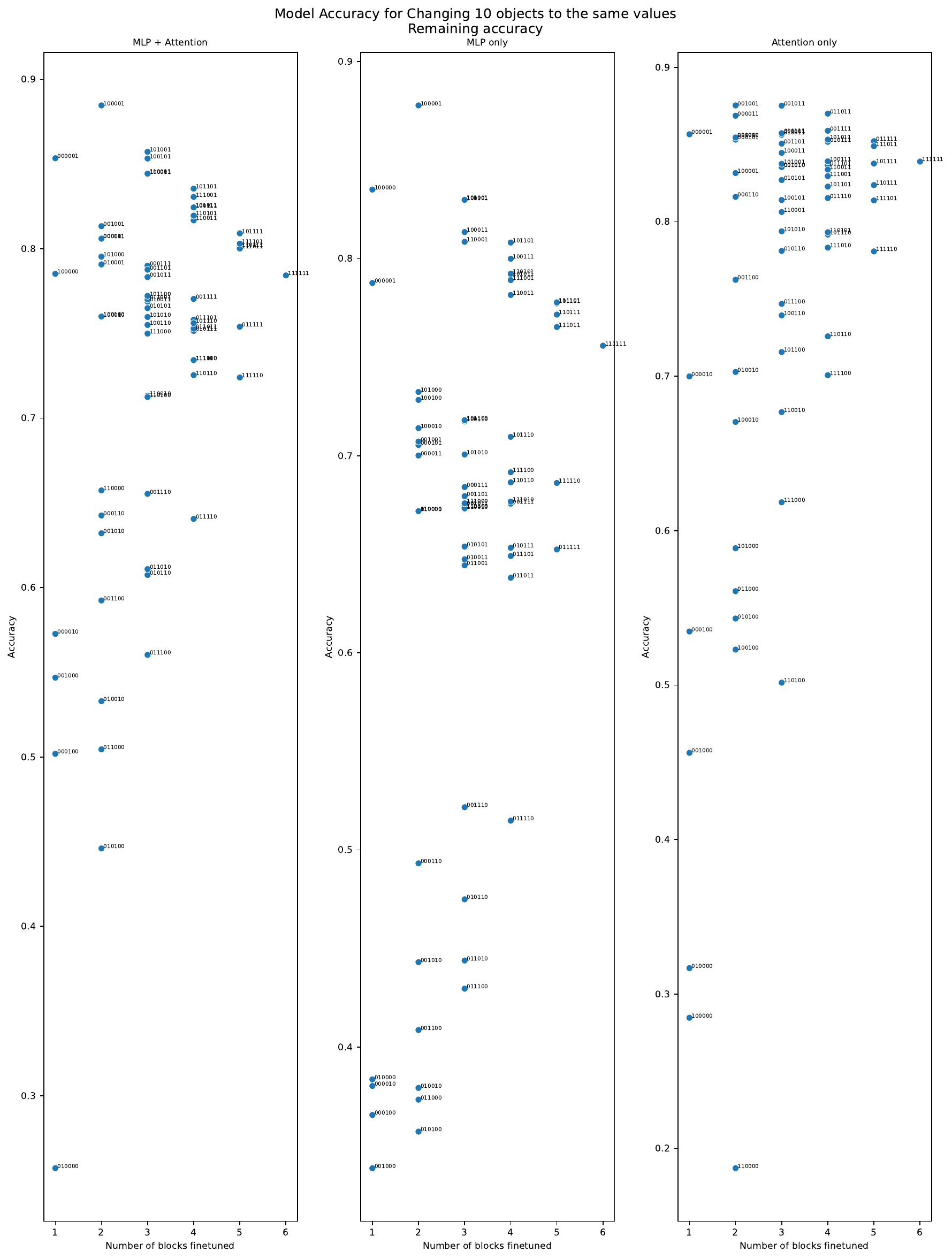}
    \caption{Correlated dataset, 10 of the same override, remaining data accuracy after full finetuning.}
    \label{fig:correlated_10overrides_allsame_remainingacc_pointplot}
\end{figure}

\begin{figure}
    \centering
    \includegraphics[width=0.95\linewidth]{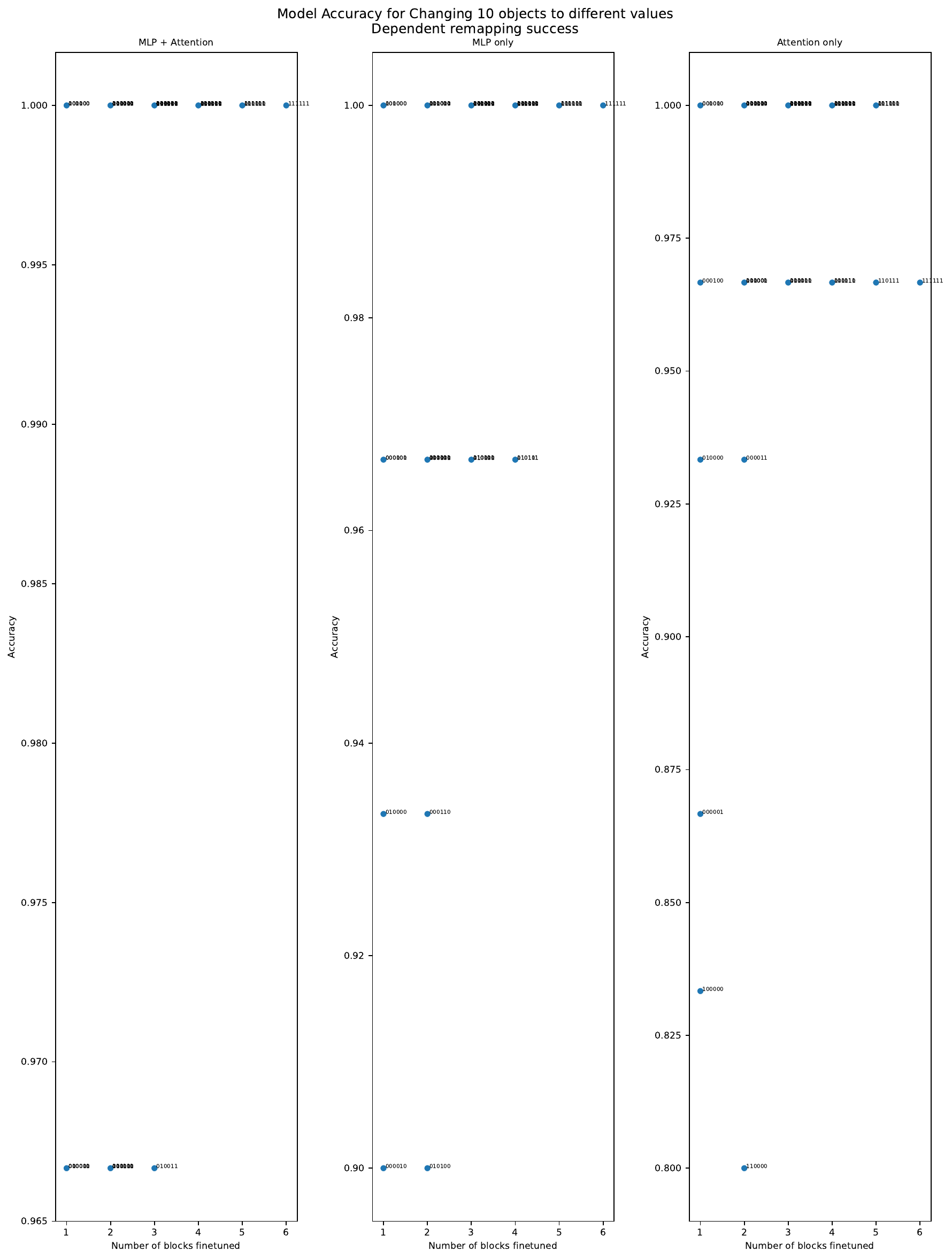}
    \caption{Correlated dataset, 10 overrides, dependent relationship editing accuracy after full finetuning.}
    \label{fig:correlated_10overrides_dependentremapping_pointplot}
\end{figure}

\begin{figure}
    \centering
    \includegraphics[width=0.95\linewidth]{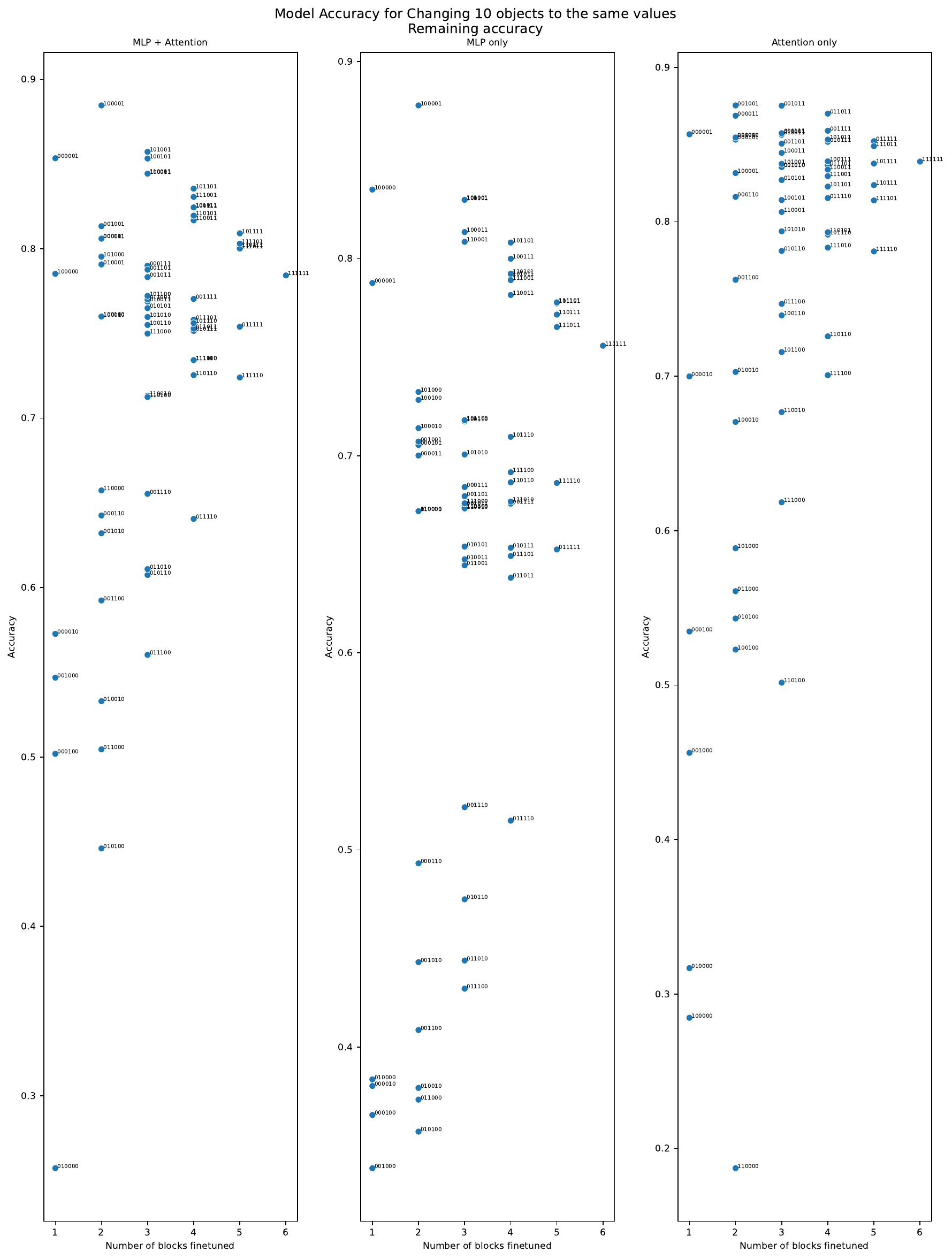}
    \caption{Correlated dataset, 10 overrides, remaining data accuracy after full finetuning.}
    \label{fig:correlated_10overrides_remainingacc_pointplot}
\end{figure}

\begin{figure}
    \centering
    \includegraphics[width=0.95\linewidth]{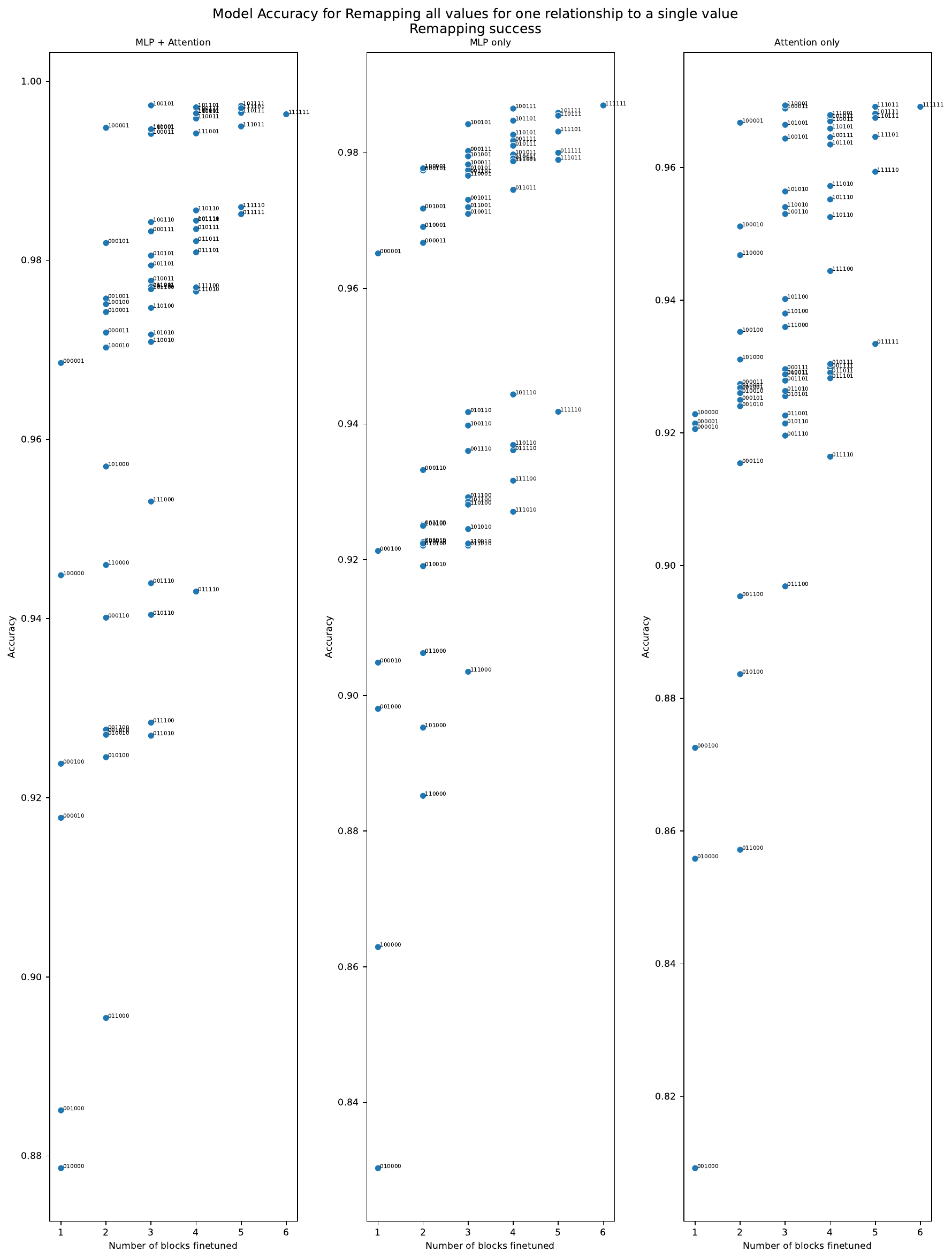}
    \caption{Correlated dataset, Forgetting R1, remaining data accuracy after full finetuning.}
    \label{fig:correlated_forgetting_remapping_pointplot}
\end{figure}

\begin{figure}
    \centering
    \includegraphics[width=0.95\linewidth]{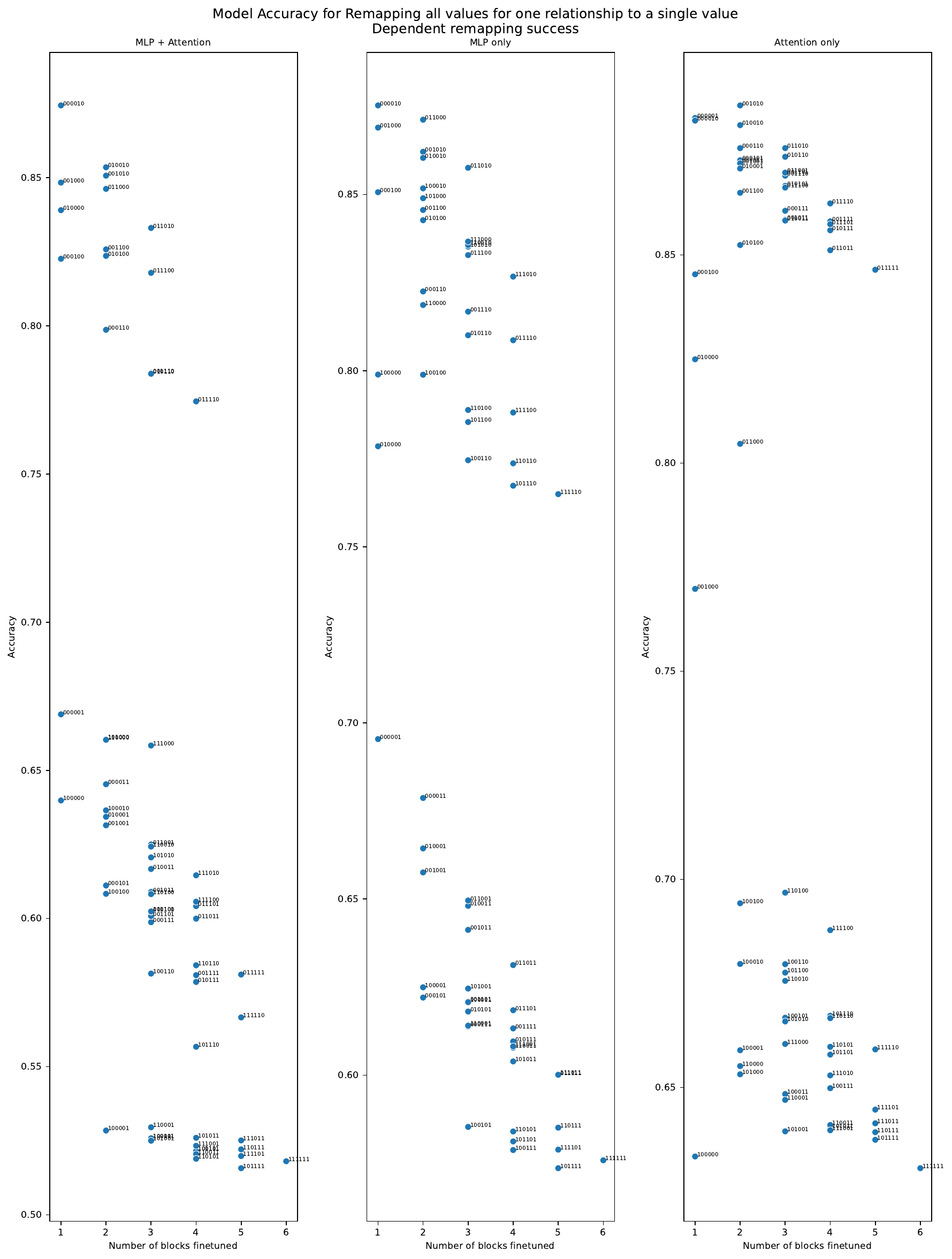}
    \caption{Correlated dataset, Forgetting R1, dependent relationship editing success after full finetuning.}
    \label{fig:correlated_forgetting_dependentremapping_pointplot}
\end{figure}

\begin{figure}
    \centering
    \includegraphics[width=0.95\linewidth]{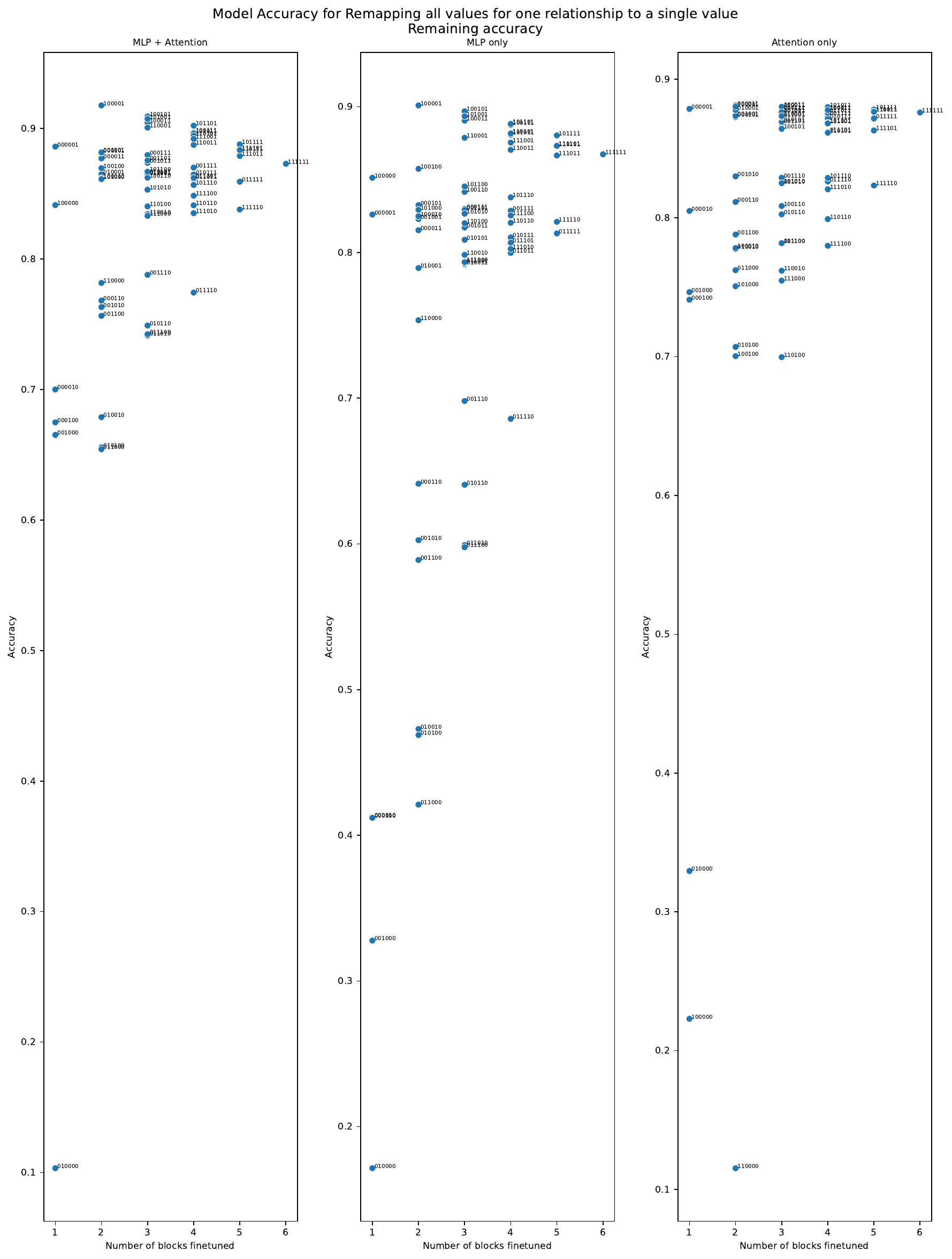}
    \caption{Correlated dataset, Forgetting R1, remaining data accuracy after full finetuning.}
    \label{fig:correlated_forgetting_remainingacc_pointplot}
\end{figure}

\end{document}